\documentclass{article}

\usepackage{microtype}
\usepackage{graphicx}
\usepackage{booktabs}
\usepackage{hyperref}

\usepackage{amsfonts}
\usepackage{mathtools}
\usepackage{amsmath,amssymb,amsthm}
\usepackage{todonotes}
\usepackage{color}
\usepackage{caption}
\usepackage{subcaption}
\usepackage[inline]{enumitem}
\usepackage[utf8]{inputenc}
\usepackage[T1]{fontenc}

\makeatletter
\def\mathcolor#1#{\@mathcolor{#1}}
\def\@mathcolor#1#2#3{
  \protect\leavevmode
  \begingroup
    \color#1{#2}#3
  \endgroup
}
\makeatother

\newcommand{\E}{\mathbb{E}}

\newcommand{\Eb}[2]{\E_{#1}\left[#2\right]}
\newcommand{\Ebb}[2]{\mathop{\E}_{#1}\left[#2\right]}

\newcommand{\cS}{\mathcal{S}}
\newcommand{\cA}{\mathcal{A}}

\newcommand{\cR}{r}
\newcommand{\cN}{\mathcal{N}}
\newcommand{\cP}{\mathcal{P}}
\newcommand{\Real}{\mathbb{R}}

\newcommand{\kl}[2]{D_{\rm KL}(#1 \ \| \ #2)}
\newcommand{\given}{|}

\DeclareMathOperator{\trace}{trace}

\newcommand\blfootnote[1]{%
  \begingroup
  \renewcommand\thefootnote{}\footnote{#1}%
  \addtocounter{footnote}{-1}%
  \endgroup
}

\pdfoutput=1

\usepackage[accepted]{icml2019}

\icmltitlerunning{Model-Based Active Exploration}

\begin{document}

\twocolumn[\icmltitle{Model-Based Active Exploration}

\begin{icmlauthorlist}
\icmlauthor{Pranav Shyam}{nn}
\icmlauthor{Wojciech Ja\'{s}kowski}{nn}
\icmlauthor{Faustino Gomez}{nn}
\end{icmlauthorlist}

\icmlaffiliation{nn}{NNAISENSE, Lugano, Switzerland}

\icmlcorrespondingauthor{Pranav Shyam}{pranav@nnaisense.com}

\icmlkeywords{Model-Based Reinforcement Learning, Curiosity, Exploration, Deep Reinforcement Learning, Planning, Bayesian Reinforcement Learning}

\vskip 0.3in
]

\printAffiliationsAndNotice{}

\begin{abstract}
  Efficient exploration is an unsolved problem in Reinforcement
  Learning which is usually addressed by reactively rewarding the agent
  for fortuitously encountering novel situations. This paper
  introduces an efficient {\em active} exploration algorithm,
  Model-Based Active eXploration (MAX), which uses an ensemble of forward models to plan to observe novel events. This is carried out by optimizing agent behaviour with respect to a measure of novelty derived from the Bayesian perspective of exploration, which is estimated using the disagreement between the futures predicted by the ensemble members. We show empirically that in semi-random discrete environments where directed exploration is critical to make
  progress, MAX is at least an order of magnitude more
  efficient than strong baselines. MAX scales to
  high-dimensional continuous environments where it builds task-agnostic models that can be used for any downstream task.
\end{abstract}

\section{Introduction}
\label{intro}


Efficient exploration in large, high-dimensional environments 
is an unsolved problem in Reinforcement Learning (RL). Current exploration
methods~\citep{osband2016deep, bellemare2016unifying,
  houthooft2016vime, pathakICMl17curiosity} are 
\textit{reactive}: the agent accidentally observes something ``novel''
and then decides to obtain more information about it.  Further
exploration in the vicinity of the novel state is carried out
typically through an exploration bonus or intrinsic motivation reward,
which have to be unlearned once the novelty has worn off, making
exploration inefficient
--- a problem we refer to as {\em
  over-commitment}.

However, exploration can also be \textit{active}, where the agent seeks
out novelty based on its own ``internal'' estimate of what action
sequences will lead to interesting transitions. This approach is inherently
more powerful than reactive exploration, but requires a method to predict the
consequences of actions and their degree of novelty. This
problem can be formulated optimally in the Bayesian setting where the
novelty of a given state transition can be measured by the
disagreement between the next-state predictions made by probable models of the environment.  

This paper introduces Model-Based Active eXploration (MAX), an
efficient algorithm, based on this principle, that approximates the
idealized distribution using an ensemble of learned forward dynamics
models.  The algorithm identifies learnable unknowns or
\textit{uncertainty} representing novelty in the environment by
measuring the amount of conflict between the predictions of the
constituent models. It then constructs exploration policies to resolve those conflicts by visiting the relevant area. Unlearnable unknowns or \textit{risk}, such as random noise in the environment does not interfere with the process since noise manifests as confusion among all models and not as a conflict. 

In discrete
environments, novelty can be evaluated using the Jensen-Shannon
Divergence (JSD) between the predicted next state distributions of the
models in the ensemble. In continuous environments, computing JSD
is intractable, so MAX instead uses the functionally equivalent Jensen-R\'enyi
Divergence based on R\'enyi quadratic entropy~\citep{renyi1961measures}.


While MAX can be used in conjunction with conventional policy learning to maximize external reward, this paper focuses on \textit{pure exploration}: exploration disregarding or in the absence of external reward, followed by exploitation~\citep{bubeck2009pure}. This setup is more natural in situations where it is useful to do task-agnostic exploration and learn models that can later be exploited for multiple tasks, including those that are not known \textit{a priori}.\blfootnote{code: \url{https://github.com/nnaisense/max}}

Experiments in the discrete domain show that MAX is significantly more efficient than reactive exploration techniques which use exploration bonuses or posterior sampling, while also strongly suggesting that MAX copes with risk. In the high-dimensional continuous Ant Maze environment, MAX reaches the far end of a U-shaped maze in just 40 episodes (12k steps), while reactive baselines are only around the mid-way point after the same time. In the Half Cheetah environment, the data collected by MAX leads to superior performance versus the data collected by reactive baselines when exploited using model-based RL. 


\section{Model-Based Active Exploration}
\label{sec:main}
The key idea behind our approach to active exploration in the environment, or the {\em
  external} Markov Decision Process (MDP), is to use a surrogate or {\em exploration} MDP where the
novelty of transitions can be estimated {\em
  before} they have actually been encountered by the agent in the
environment.  The next section provides the formal context for the
conceptual foundation of this work.

\subsection{Problem Setup}
Consider the environment or the external MDP,
represented as the tuple $(\cS, \cA, t^*, \cR, \rho_0)$, where $\cS$ is the state
space, $\cA$ is the action space,
$t^*$ is the unknown transition function, $\cS \times \cA \times \cS \rightarrow [0, \infty)$,  specifying the probability density $p(s' \given s, a, t^*)$ of the next state $s'$ given the current state $s$ and the action $a$,
$\cR: \cS \times \cA \rightarrow \Real$ is the reward function,
$\rho_0: \cS \rightarrow [0, \infty)$ is the probability density function of the initial
state.

Let $T$ be the space of all possible transition functions and
$\cP(T)$ be a probability distribution over transition functions
that captures the current belief of how the environment works, with corresponding density function $p(T)$.

The objective of pure exploration is to efficiently accumulate information about the environment, irrespective of $\cR$. This is equivalent to learning an accurate model of the transition function, $t^*$, while minimizing the number of state transitions, $\phi$, belonging to transition space $\Phi$, required to do so, where $\phi = (s, a, s')$ and $s'$ is the state resulting from action $a$ being taken in state $s$.

Pure exploration can be defined as an iterative process, where in each iteration, an exploration policy $\pi: \cS \times \cA \rightarrow [0, \infty)$, specifying a density $p(a|s,\pi)$, is used to collect information about areas of the environment that have not been explored up to that iteration. To learn such exploration policies, there needs to be a method to evaluate any given policy at each iteration.


\subsection{Utility of an Exploration Policy}\label{sec:util-an-expl}
In the standard RL setting, a policy would be learned to take actions
that maximize some function of the external reward received from the
environment according to $\cR$, i.e., the~{\em return}. Because pure
active exploration does not care about $\cR$, and $t^*$ is unknown, the amount of new information conveyed about
the environment by state transitions that {\em could} be caused by an exploration policy has to be used as the learning signal.

From the Bayesian perspective, this can be captured by the KL-divergence
between $\cP(T)$, the (prior) distribution over transition
functions before a particular transition $\phi$, and $\cP(T \given \phi)$, the posterior distribution after $\phi$ has occurred. This is
commonly referred to as Information Gain, which is abbreviated as  $\mathrm{IG}(\phi)$ for a transition $\phi$:
\begin{equation}
    \label{eq:kl}
    \mathrm{IG}(s, a, s') = \mathrm{IG}(\phi) = \kl{\cP(T \given  \phi)}{\cP(T
      )}.
\end{equation}

The utility can be understood as the extra number of bits needed to
specify the posterior relative to the prior, effectively, the number of
bits of information that was gathered about the external MDP.
Given $\mathrm{IG}(\phi)$, it is now possible to compute the utility of the exploration policy,
$\mathrm{IG}(\pi)$, which is the expected utility over the transitions
when $\pi$ is used:
\begin{equation}
    \label{eq:utility_pi_initial}
    \mathrm{IG}(\pi) = \Eb{\phi \sim \cP(\Phi \given \pi)}{\mathrm{IG}(\phi)},
\end{equation}

\noindent which can be expanded into  (see Appendix~\ref{sec:appendix_policy_eval}):
\begin{equation}
    \label{eq:policy-valu-pre}
    \mathrm{IG}(\pi) = \Eb{t \sim \cP(T)}{\Eb{s, a \sim \cP(\cS, \cA \given  \pi, t)} {u(s, a)}},
\end{equation}
where
\begin{equation}
    \label{eq:utility_state_action_margin}
    u(s, a) = \int_T \int_{\cS} \mathrm{IG}(s, a, s') p(s' \given  a, s, t) p(t)\, ds' dt.
\end{equation}

It turns out that (see Appendix ~\ref{sec:appendix_action_eval}):
\begin{equation}
    \label{eq:final_closed_form}
    u(s, a) = \mathrm{JSD}\{\cP(\cS \given  s, a, t)  \, \mid \, t \,  \sim  \, \cP(T) \} ,
\end{equation}
where $\mathrm{JSD}$ is the Jensen-Shannon Divergence,
which captures the amount of disagreement present in a space of
distributions. Hence, the \textit{utility} of the state-action pair
$u(s, a)$ is the disagreement, in terms of JSD, among
the next-state distributions given $s$ and $a$ of all possible transition functions weighted by their probability. Since
$u$ depends only on the prior $\cP(T)$, the novelty of potential transitions can be calculated without having to actually effect them in the external MDP. 

The ``internal'' exploration MDP can then be defined as $(\cS, \cA, \bar{t}, u, \delta(s_{\tau}))$, where the sets $\cS$ and $\cA$ are
identical to those of the external MDP, and the
transition function $\bar{t}$ is defined such that,
\begin{equation}
    p(s' \given s, a, \bar{t}) := \Ebb{t \sim \cP(T)}{p(s'\given s, a, t)},
\end{equation}
which can be interpreted as a different sample of
$\cP(T)$ drawn at each state transition. $u$ is to the exploration MDP what the $\cR$
is to the external MDP, so that maximizing $u$ results in the
optimal exploration policy at that iteration, just as
maximizing $\cR$ results in the optimal policy for the corresponding task. Finally, the initial state
distribution density is set to the Dirac delta function $\delta(s_{\tau})$ such that the initial
state of exploration MDP is always the current state $s_{\tau}$ in the environment. It is important to understand that the prior $\cP(T)$ is used twice in the exploration MDP:
\begin{enumerate}
    \item To specify the state-action joint distribution as per Equation~\ref{eq:policy-valu-pre}. Each member $t$ in the prior $\cP(T)$ determines a distribution $\cP(\cS, \cA \given \pi, t)$ over the set of possible state-action pairs that can result by sequentially executing actions according to $\pi$ starting in $s_{\tau}$.
    \item To obtain the utility for a particular transition as per Equation~\ref{eq:final_closed_form}.  Each
      state-action pair $(s, a)$ in the above $\cP(\cS, \cA \given \pi, t)$, according to the transition functions from $\cP(T)$, forms a set of
      predicted next-state distributions $\{\cP(\cS \given s, a, t) \, \mid \, t \, \sim \, \cP(T)\}$. The JSD of this set is $u(s,a)$.
\end{enumerate}



\subsection{Bootstrap Ensemble Approximation}
\label{sec:bootstr-ensemble-app}

The prior $\cP(T)$ can be approximated using a bootstrap
ensemble of $N$ learned transition functions or models that are each
trained independently using different subsets of the {\em history} $D$, consisting of state transitions experienced by the agent while
exploring the external MDP~\citep{efr13bay}.  Therefore, while
$\cP(T)$ is uniform when the agent is initialized, thereafter it
is conditioned on agent's history $D$ so that the general form
of the prior is $\cP(T \given D)$.  For generalizations
that are warranted by observed data, there is a good chance that the
models make similar predictions. If the generalization is
not warranted by observed data, then the models could disagree owing
to their exposure to different parts of the data distribution.

Since the ensemble contains models that were trained to accurately
approximate the data, these models have higher probability densities than a random model.
Hence, even a relatively small ensemble can approximate the true distribution,
$\cP(T \given D)$~\citep{lakshminarayanan2017simple}. Recent work suggests that it is possible to do so even in high-dimensional state and action spaces~\citep{kurutach2018model, chua2018deep}. Using an $N$-model ensemble $\{t_1, t_2, \cdots, t_N\}$ approximating the prior and assuming that all models fit the data equally well, the dynamics of the exploration MDP can be
approximated by randomly selecting one of the $N$ models with equal probability at each transition.

Therefor, to approximate $u(s,a)$, the JSD in
Equation~\ref{eq:final_closed_form} can be expanded as (see Appendix
~\ref{sec:appendix_action_eval}):
\begin{align}
    \begin{split}
    \label{eq:final_closed_form_entropy}
    u(s, a) &= \mathrm{JSD}\{\cP(\cS \given  s, a, t)  \, \mid \, t \,  \sim  \, \cP(T) \}\\
    &= H\left(\Eb{t \sim \cP(T)}{ \cP(\cS \given  s, a, t)} \right) \\
    & \quad \quad - \Eb{t \sim \cP(T)}{ H(\cP(\cS \given  s, a, t))} ,
    \end{split}
\end{align}

where $H(\cdot)$ denotes the entropy of the
distribution. Equation~\ref{eq:final_closed_form_entropy} can be summarized as the difference between
the entropy of the average and the average entropy and
can be approximated by averaging samples from the ensemble:
\begin{equation}
    \begin{split}
    \label{eq:final-policy-utility}
    u(s, a) & \simeq H\left(\frac{1}{N}\sum_{i=1}^N \cP(\cS \given  s, a, t_i) \right) \\
    & \quad \quad - \frac{1}{N} \sum_{i=1}^N H(\cP(\cS \given  s, a, t_i)) .
    \end{split}
\end{equation}

\subsection{Large Continuous State Spaces}
\label{sec:cont_renyi}
For continuous state spaces, $\cS \subset \Real^{d}$, the next-state distribution $\cP(\cS \given  s, a, \tilde{t}_i)$ is generally parameterized, typically, using a multivariate Gaussian $\mathcal{N}_i(\mu_i, \Sigma_i)$ with mean vector $\mu_i$ and co-variance matrix $\Sigma_i$. With this, evaluating Equation~\ref{eq:final-policy-utility} is intractable as it involves estimating the entropy of a mixture of Gaussians, which has no analytical solution. This problem can be circumvented by replacing the Shannon entropy with R\'enyi entropy~\citep{renyi1961measures} and using the corresponding Jensen-R\'enyi Divergence (JRD).

The R\'enyi entropy of a random variable $X$ is defined as
\[
H_{\alpha}(X) = \frac{1}{1 - \alpha} \ln \int p(x)^{\alpha} dx 
\]
for a given order $\alpha \ge 0$, of which Shannon entropy is a special case when $\alpha $ tends to $1$. When $\alpha=2$, the resulting quadratic R\'enyi entropy has a closed-form solution for a mixture of $N$ Gaussians~\citep{wang2009closed}. Therefore, the Jensen-R\'enyi Divergence $\mathrm{JRD}\{\cN_i(\mu_i, \Sigma_i) \mid i = 1, \dots , N\}$ is given by
\begin{equation*}
    H_{2}\left(\sum_i^N \frac{1}{N}\cN_i\right) - \frac{1}{N} \sum_i^N H_{2}\left(\cN_i\right) ,
\end{equation*}
can be calculated with
\begin{equation}
\label{eq:final-renyi}
-\ln \left[\frac{1}{N^2}\sum_{i, j}^{N} \mathfrak{D}(\cN_i, \cN_j) \right] - \frac{1}{N} \sum_i^N  \frac{\ln|\Sigma_i|}{2} - c ,
\end{equation}
where $c = d \ln(2) / 2$ and
\begin{equation}
\mathfrak{D}(\cN_i, \cN_j) = \frac{1}{\left| \Omega \right|^{\frac{1}{2}}} \exp{\left(-\frac{1}{2}\Delta^{T}\Omega^{-1}\Delta\right)} ,
\end{equation}
with $\Omega = \Sigma_i+\Sigma_j$ and $\Delta = \mu_j - \mu_i$.

\begin{figure*}
    \centering
    \begin{subfigure}[b]{\textwidth}
        \centering
        \includegraphics[width=0.9\columnwidth]{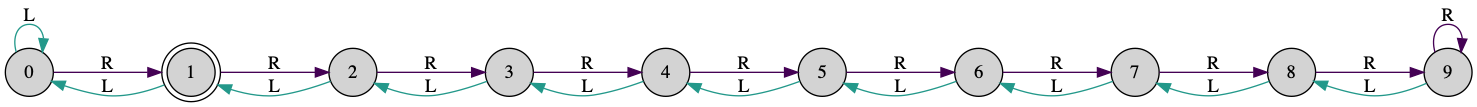}
        \caption{Chain environment of length $10$. }
        \label{fig:chain_env}
    \end{subfigure}
    \vskip 0.2in
    \begin{subfigure}[b]{0.325\textwidth}
        \centering
        \includegraphics[width=\columnwidth]{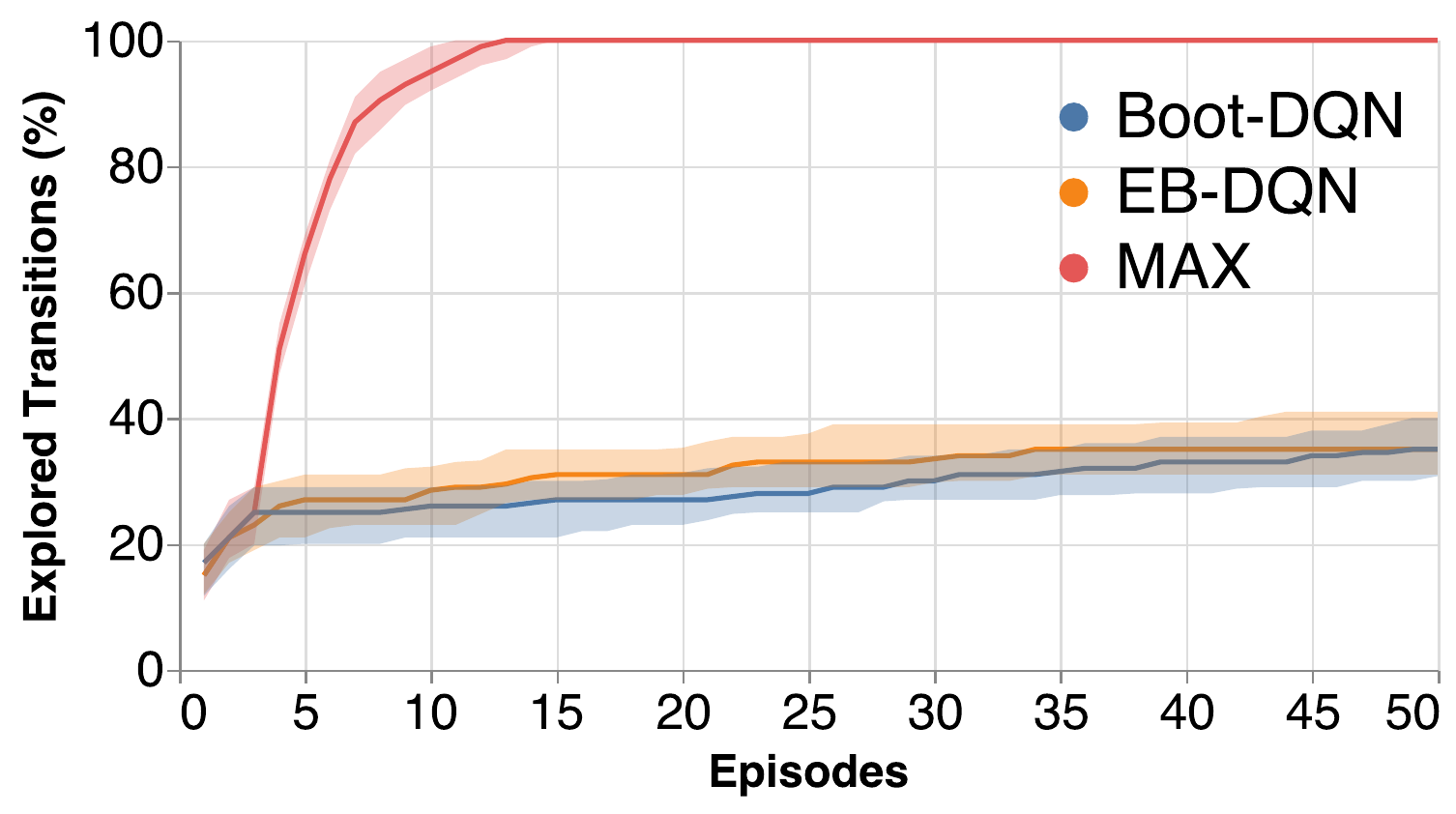}
        \caption{$50$-state chain.}
        \label{fig:main_focus}
    \end{subfigure}
    \hfill
    \begin{subfigure}[b]{0.325\textwidth}
        \centering
        \includegraphics[width=\columnwidth]{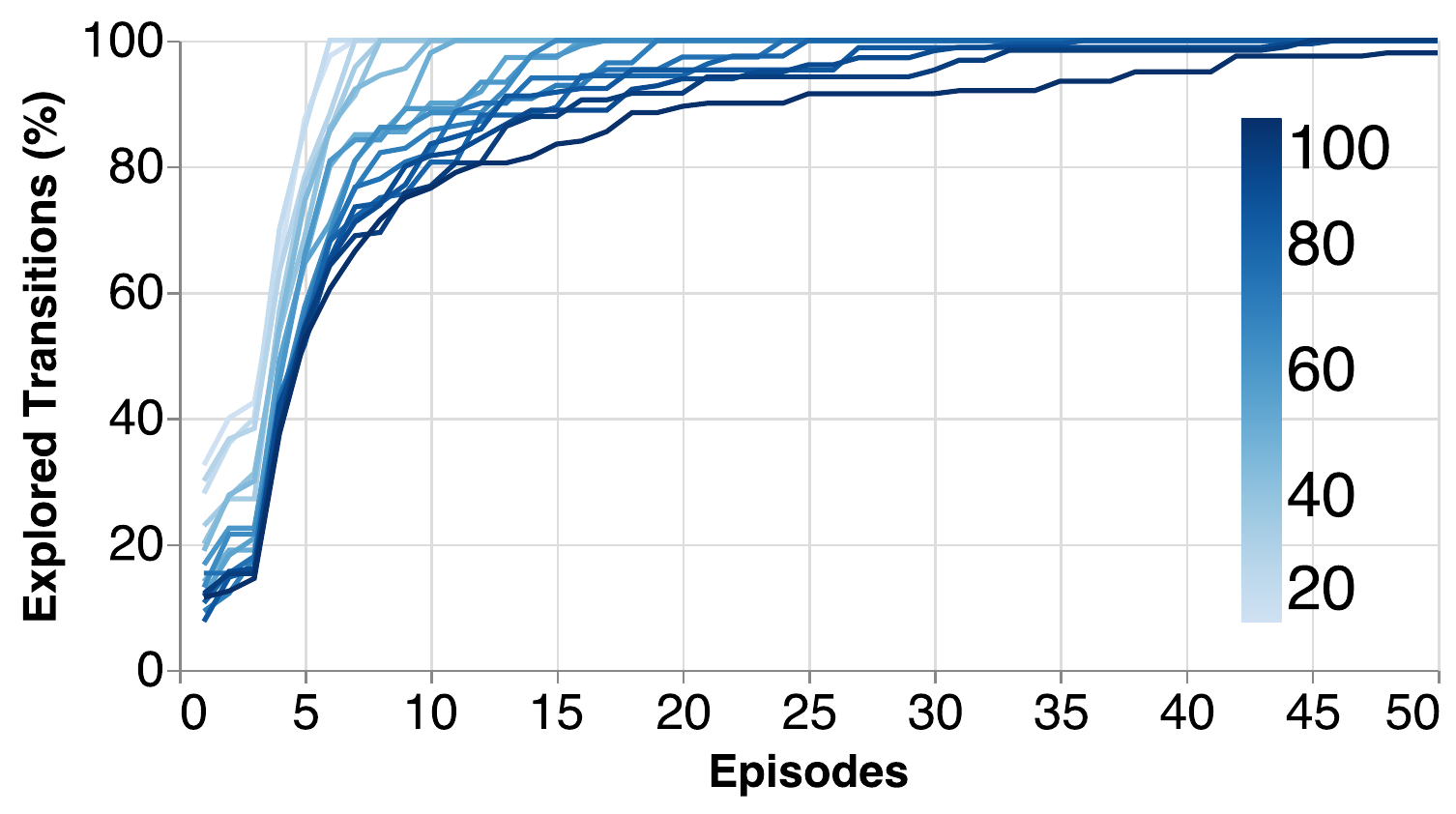}
        \caption{Chain lengths}
        \label{fig:chain_length_ablation}
    \end{subfigure}
    \hfill
    \begin{subfigure}[b]{0.325\textwidth}
        \centering
        \includegraphics[width=\columnwidth]{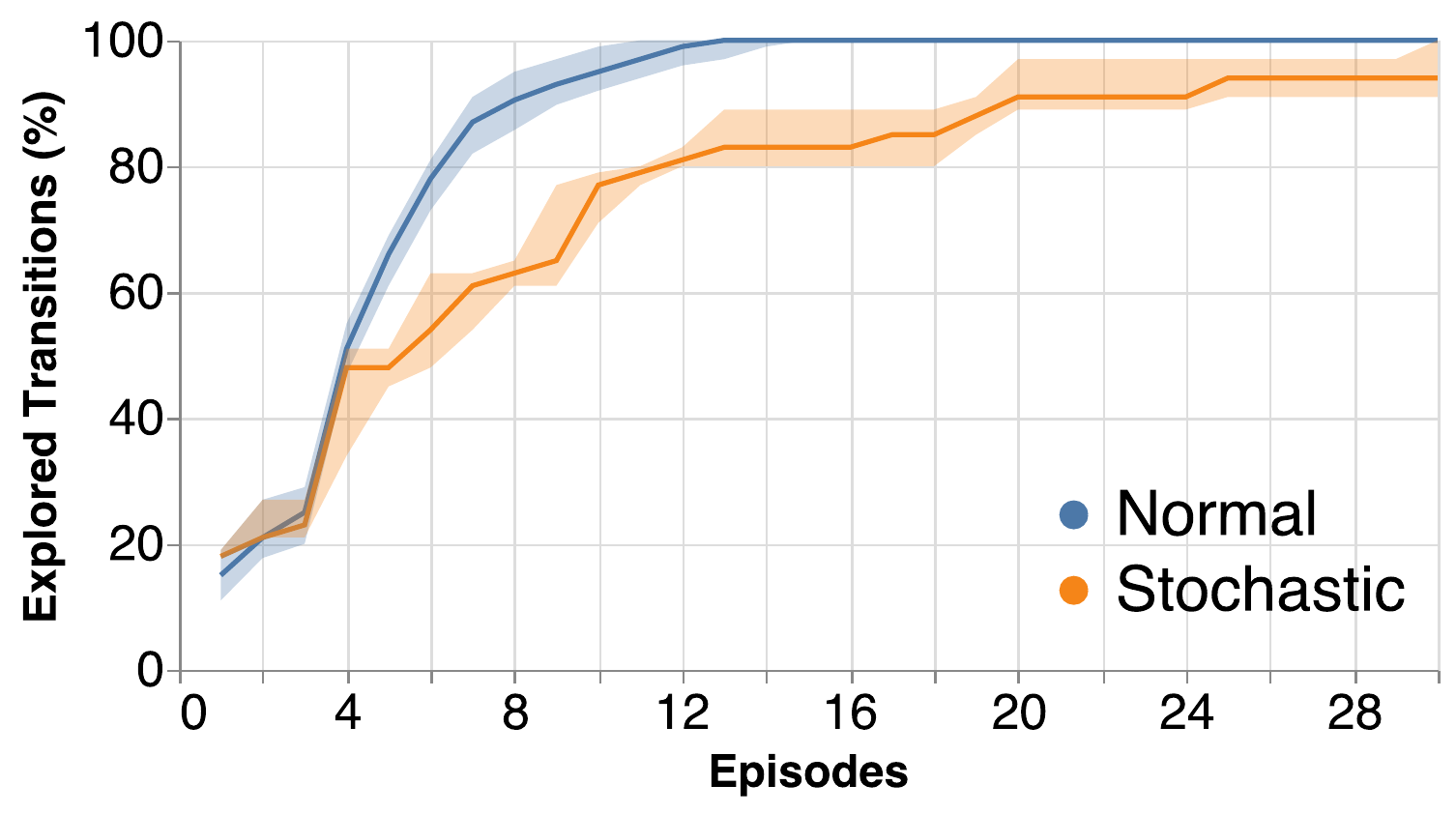}
        \caption{Stochastic trap}
        \label{fig:stochastic_ablation}
    \end{subfigure}
    \caption{{\bf Algorithm performance on the randomized Chain environment}. For the first $3$ episodes, marked by the vertical dotted line, actions were chosen at random (as warm-up). Each line corresponds to the median of $100$ runs (seeds) in (\subref{fig:main_focus}) and $5$ runs in (\subref{fig:chain_length_ablation}) and (\subref{fig:stochastic_ablation}). The shaded area spans the $25$th and $75$th percentiles.}
\end{figure*}

Equation~\ref{eq:final-renyi} measures divergence among predictions of models based on some combination of their means and variances. However, when the models are learned, the parameters concentrate around their true values at different rates and the environments can greatly differ in the amount of noise they contain. On the one hand, if the environment is completely deterministic, exploration effort could be wasted in precisely matching the small predicted variances of all the models. On the other hand, ignoring the variance in a noisy environment could result in poor exploration. To inject such prior knowledge into the system, an optional temperature parameter $\lambda \in [0, 1]$ that modulates the sensitivity of Equation~\ref{eq:final-renyi} with respect to the variances was introduced. Since the outputs of parametric non-linear models, such as neural networks, are unbounded, it is common to use variance bounds for model and numerical stability. Using the upper bound $\Sigma_{U}$ the variances can be re-scaled with $\lambda$:
\begin{equation*}
    \hat{\Sigma}_i = \Sigma_{U} - \lambda (\Sigma_{U} - \Sigma_i) \quad \forall \: i=1,\dots,N .
\end{equation*}

In this paper, $\lambda$ was fixed to $0.1$ for all continuous environments.

\begin{algorithm}[t]
    \caption{\sc Model-Based Active Exploration}
    \label{alg:main}
    \begin{algorithmic}
        \STATE {\textbf{Initialize:} Transitions dataset $D$, with random policy}
        \STATE {\textbf{Initialize:} Model ensemble, $\tilde{T} = \{t_1, t_2, \cdots, t_N\}$}
        \REPEAT
            \WHILE {episode not complete}
                \STATE{ExplorationMDP $\leftarrow (\cS, \cA, \mathrm{Uniform}\{\tilde{T}\}, u, \delta(s_{\tau}))$}
                \STATE{$\pi \leftarrow $ {\sc Solve}(ExplorationMDP)}
                \STATE {$a_{\tau} \sim \pi(s_{\tau})$}
                \STATE {act in environment: $s_{\tau+1} \sim \cP(\cS| s_{\tau},a_{\tau},t^*)$}
                \STATE {$D \leftarrow D \cup \{(s_{\tau}, a_{\tau}, s_{\tau+1})\}$} 
                \STATE {Train $t_i$ on $D$ for each $t_i$ \textbf{in} $\tilde{T}$}
            \ENDWHILE
        \UNTIL {computation budget exhausted}
    \end{algorithmic}
\end{algorithm}

\subsection{The MAX Algorithm}
\label{sec:algorithm}
Algorithm~\ref{alg:main} presents MAX in high-level pseudo-code. MAX is, essentially, a model-based RL algorithm with exploration as
its objective. At each step, a fresh exploration policy is learned to maximise its return in the exploration MDP, a procedure which is generically specified as $\textsc{Solve}(\text{ExplorationMDP})$. The policy then acts in the external MDP to collect new data, which is used to train the ensemble yielding the
approximate posterior. This posterior is then used as the approximate prior in the subsequent exploration step. Note that a transition function is drawn from $\tilde{T}$ for each transition in the exploration MDP. In practice, training the model ensemble and optimizing the policy can be performed at a fixed frequency to reduce the computational cost.

\begin{figure*}[htb]
    \vskip 0.2in
    \centering
    \hfill
    \begin{subfigure}[b]{0.25\textwidth}
        \centering
        \includegraphics[width=\columnwidth]{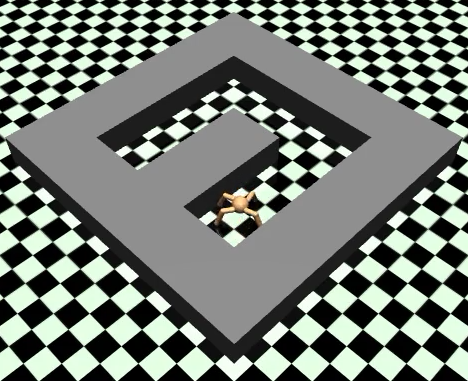}
        \vspace{0.1in}
        \caption{Ant Maze Environment}
        \label{fig:mujoco_maze}
    \end{subfigure}
    \hfill
    \begin{subfigure}[b]{0.725\textwidth}
        \centering
        \includegraphics[width=0.95\columnwidth]{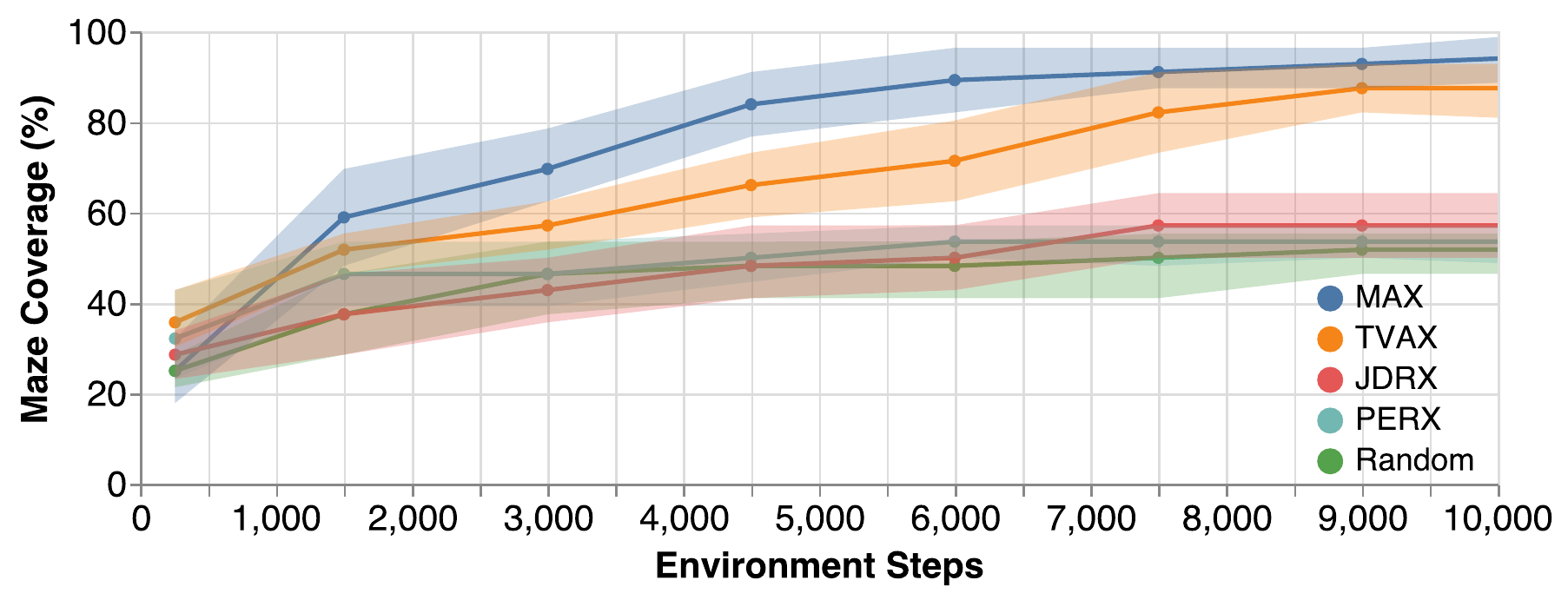}
        \caption{Maze Exploration Performance}
        \label{fig:maze_perf}
    \end{subfigure}
    \vskip 0.1in
    \begin{subfigure}[b]{0.24\textwidth}
        \centering
        \includegraphics[height=0.75\columnwidth]{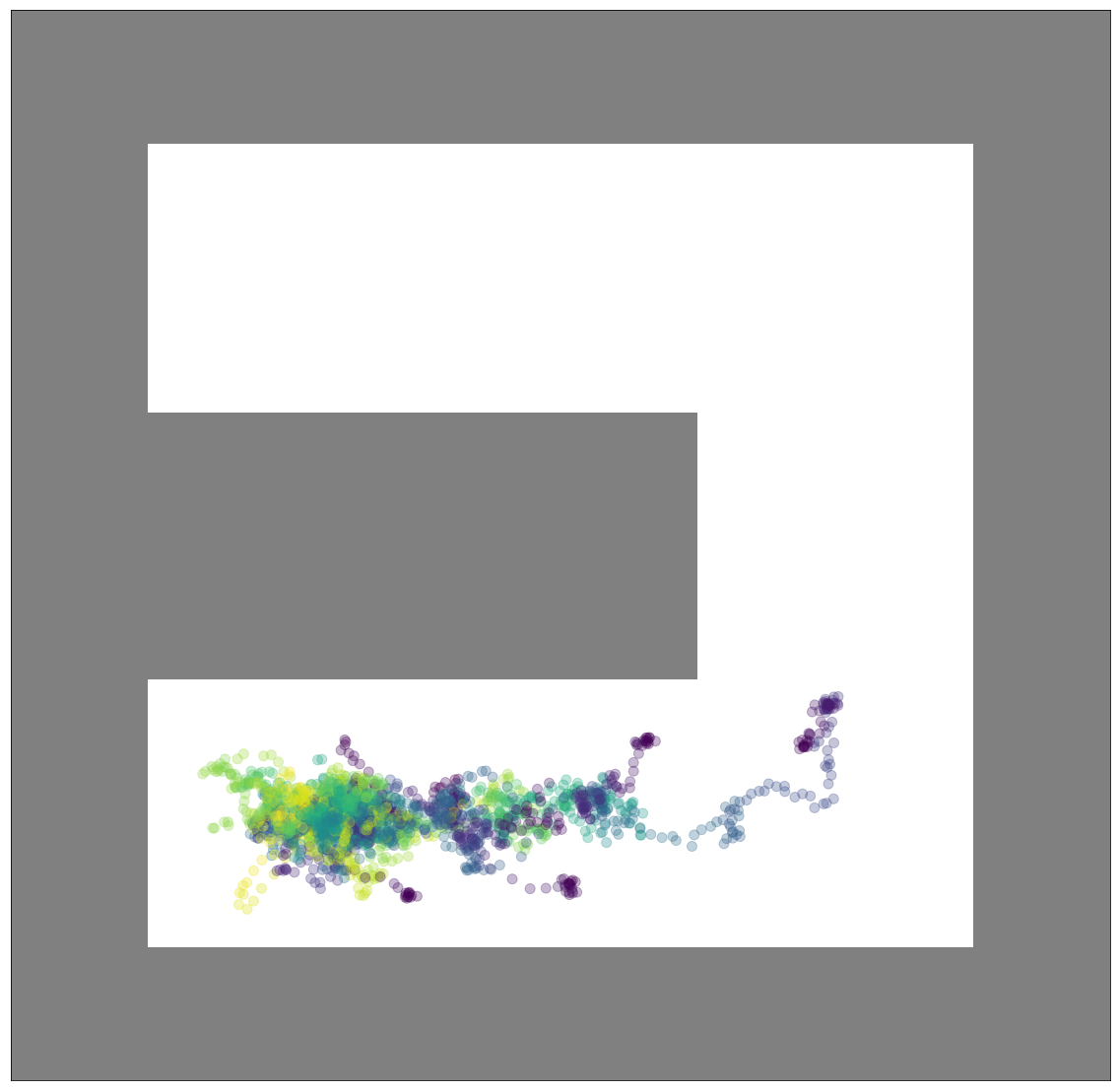}
        \caption{300 steps}
        \label{fig:maze_viz_step_300}
    \end{subfigure}
    \hfill
    \begin{subfigure}[b]{0.24\textwidth}
        \centering
        \includegraphics[height=0.75\columnwidth]{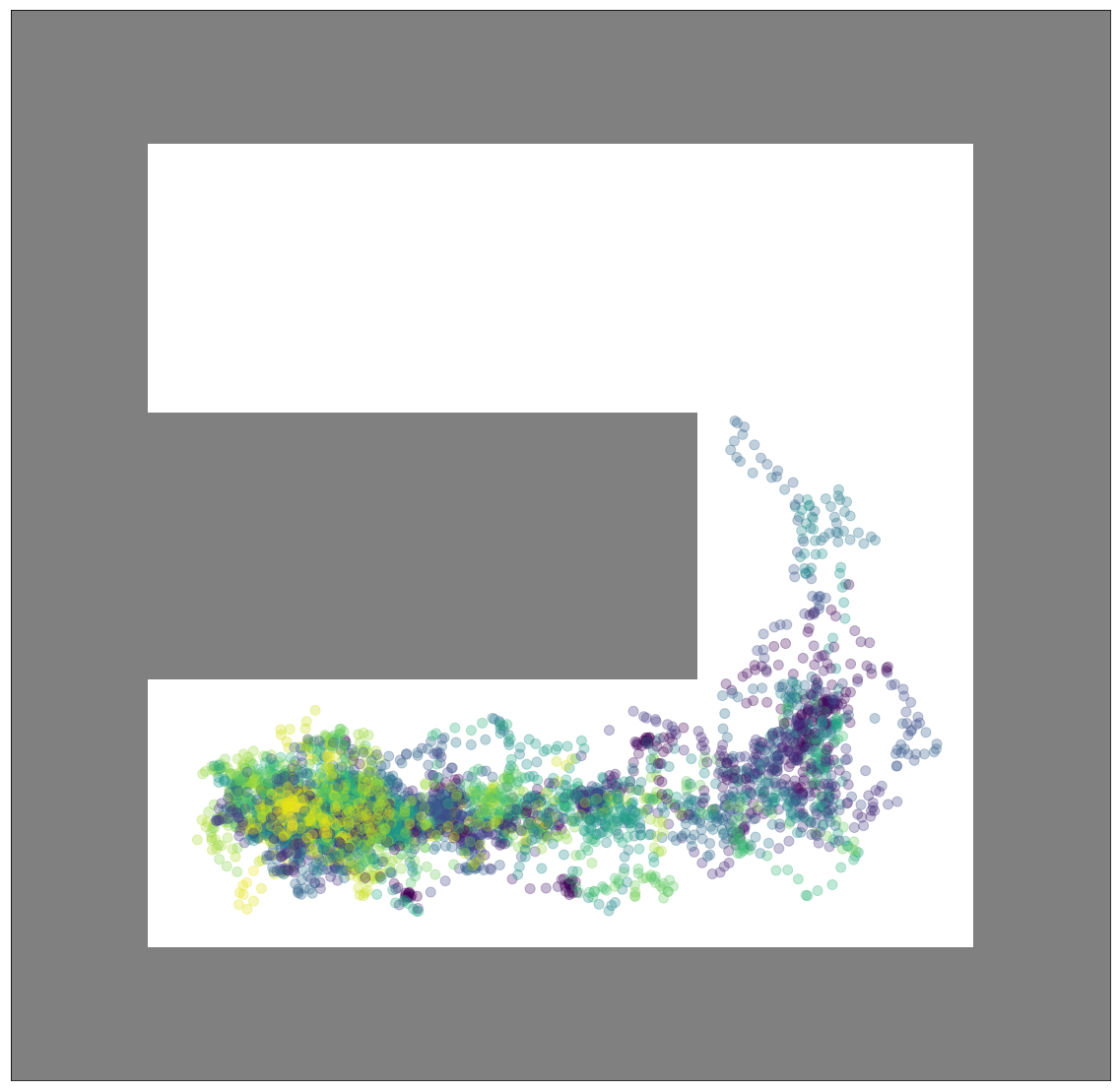}
        \caption{600 steps}
        \label{fig:maze_viz_step_600}
    \end{subfigure}
    \hfill
    \begin{subfigure}[b]{0.24\textwidth}
        \centering
        \includegraphics[height=0.75\columnwidth]{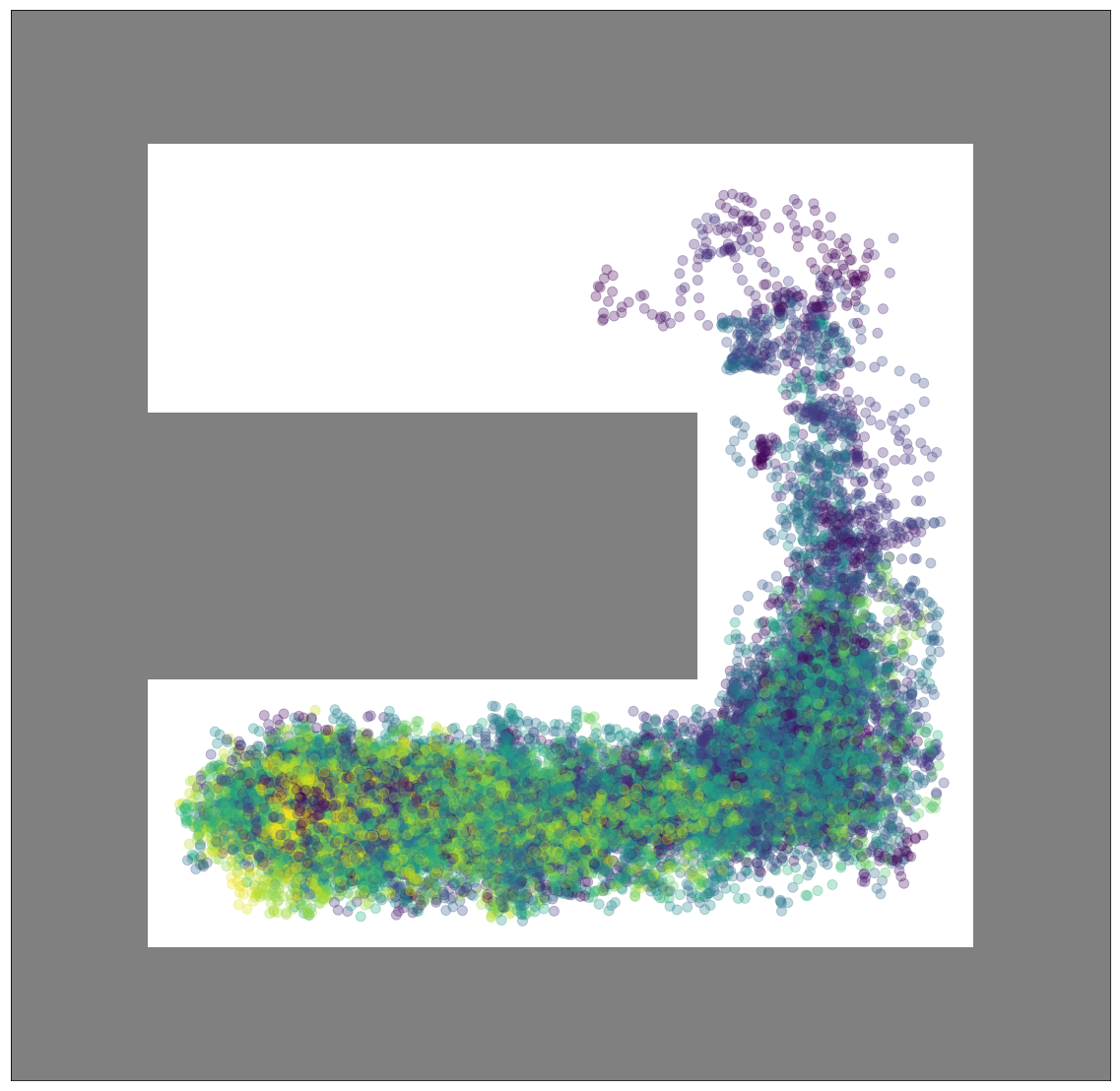}
        \caption{3600 steps}
        \label{fig:maze_viz_step_3600}
    \end{subfigure}
    \hfill
    \begin{subfigure}[b]{0.24\textwidth}
        \centering
        \includegraphics[height=0.75\columnwidth]{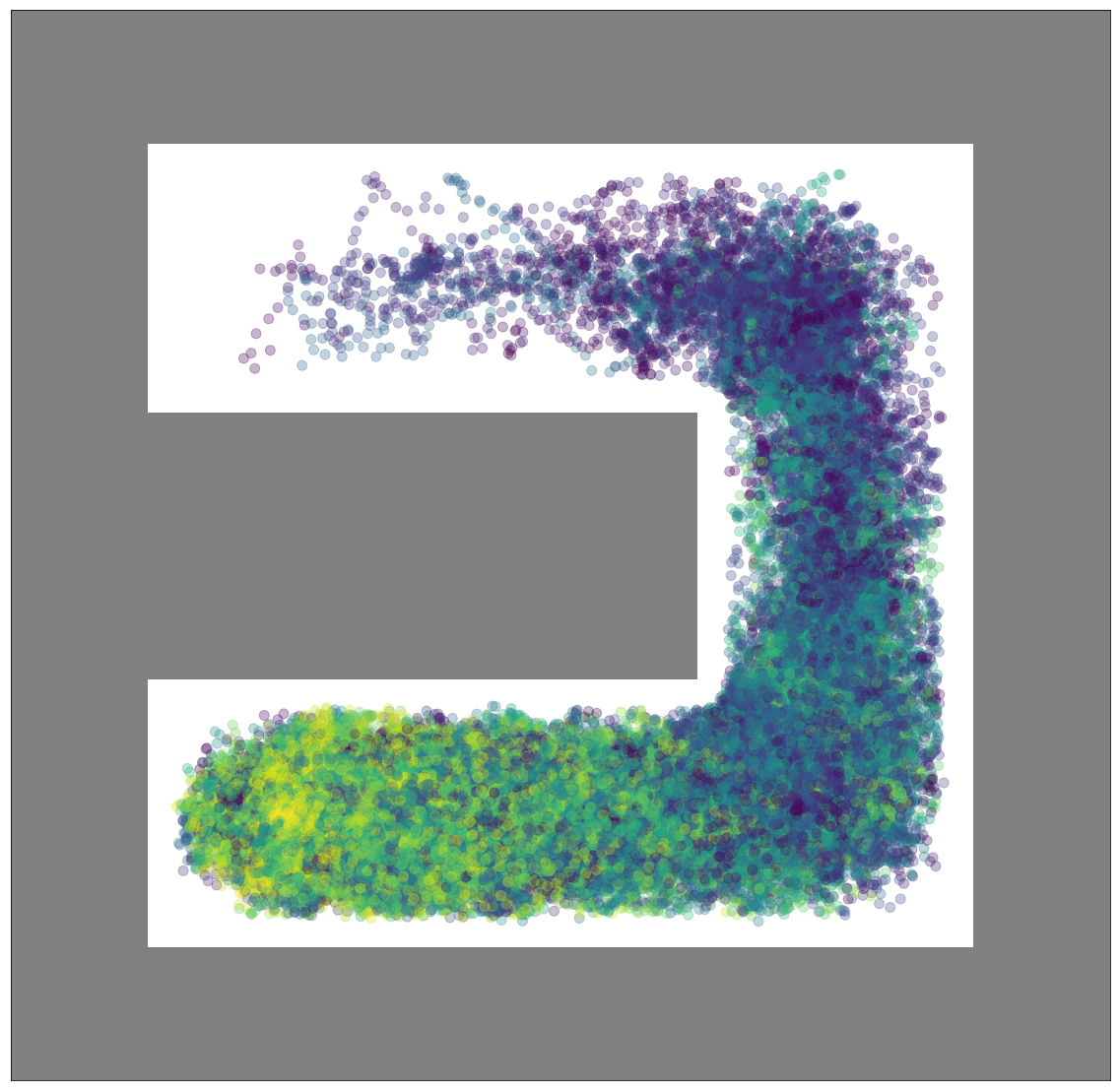}
        \caption{12000 steps}
        \label{fig:maze_viz_step_12000}
    \end{subfigure}
    \caption{{\bf Performance of MAX exploration on the Ant Maze task}. (\subref{fig:mujoco_maze}) shows the environment used. Results presented in (\subref{fig:maze_perf}) show that active methods (MAX and TVAX) are significantly quicker in exploring the maze compared to reactive methods (JDRX and PREX) with MAX being the quickest. (\subref{fig:maze_viz_step_300})-(\subref{fig:maze_viz_step_12000}) visualize the maze exploration by MAX across 8 runs. Chronological order of positions within an episode is encoded with the color spectrum, going from yellow (earlier in the episode) to blue (later in the episode).}
    \label{fig:ant_maze_main}
\end{figure*}


\section{Experiments}
\label{sec:experiments}

\subsection{Discrete Environment}
A randomized version of the Chain environment
(Figure~\ref{fig:chain_env}), designed to be hard to explore proposed
by~\citet{osband2016deep}, which is a simplified generalization of
RiverSwim~\citep{strehl2005theoretical} was used to evaluate MAX. 
Starting in the second state (state 1) of an
$L$-state chain, the agent can move either left or right at each
step. An episode lasts $L + 9$ steps, after which the agent is
reset to the start. The agent is first given $3$ warm up episodes
during which actions are chosen randomly. Trying to
move outside the chain results in staying in place. The agent is
rewarded only for staying in the edge states: $0.001$ and $1$ for the
leftmost and the rightmost state, respectively. To make the problem
harder (e.g.\ not solvable by the policy {\tt\small always-go-right}),
the effect of each action was randomly swapped so that in approximately half of the states,
going \textsc{right} results in a leftward transition and
vice-versa. Unless stated
otherwise, $L=50$ was used, so that exploring the
environment fully and reaching the far right states is unlikely using random exploration.
The probability of the latter decreases exponentially with $L$. Therefore, in
order to explore efficiently, an agent needs to exploit the structure of the environment.

MAX was compared to two exploration methods based on the \textit{optimism in face
  of uncertainty} principle~\citep{kaelbling1996reinforcement}:
Exploration Bonus DQN~(EB-DQN;~\citealp{bellemare2016unifying}) and Bootstrapped
DQN~(Boot-DQN;~\citealp{osband2016deep}). Both algorithms
employ the sample efficient DQN
algorithm~\citep{mnih2015human}. Bootstrapped DQN is claimed to be
better than ``state of the art approaches to exploration via dithering
($\epsilon$-greedy), optimism and posterior
sampling''~\citep{osband2016deep}. 
Both of them are \textit{reactive}
since they do not explicitly seek new transitions, but upon finding
one, prioritize frequenting it. Note that these baselines are fundamentally ``any-time-optimal'' RL algorithms which minimize cumulative regret by trading-off exploration and exploitation in each action. 

For the chain environment, MAX used Monte-Carlo Tree Search to find open-loop exploration policies (see Appendix~\ref{app:chain_env} for details). The hyper-parameters for both of the baseline methods were tuned with grid search.

Figure~\ref{fig:main_focus} shows the percentage of explored transitions as a function of training episodes for all the methods. MAX explores $100\%$ of the
transitions in around $15$ episodes while the baseline methods reach $40\%$ in $60$ episodes. Figure~\ref{fig:chain_length_ablation} shows the exploration progress curves for MAX when chain length was varied from $20$ to $100$ in intervals of $5$. 

To see if MAX can distinguish between the environment risk and uncertainty, the left-most state (state $0$) of the Chain
Environment was modified to be a \textit{stochastic trap} state (see Appendix~\ref{app:chain_env}). Although MAX slowed down as a consequence, it still
managed to explore the transitions as Figure~\ref{fig:stochastic_ablation} shows.

\begin{figure*}
    \centering
    \begin{subfigure}[b]{0.325\textwidth}
        \centering
        \includegraphics[width=\textwidth]{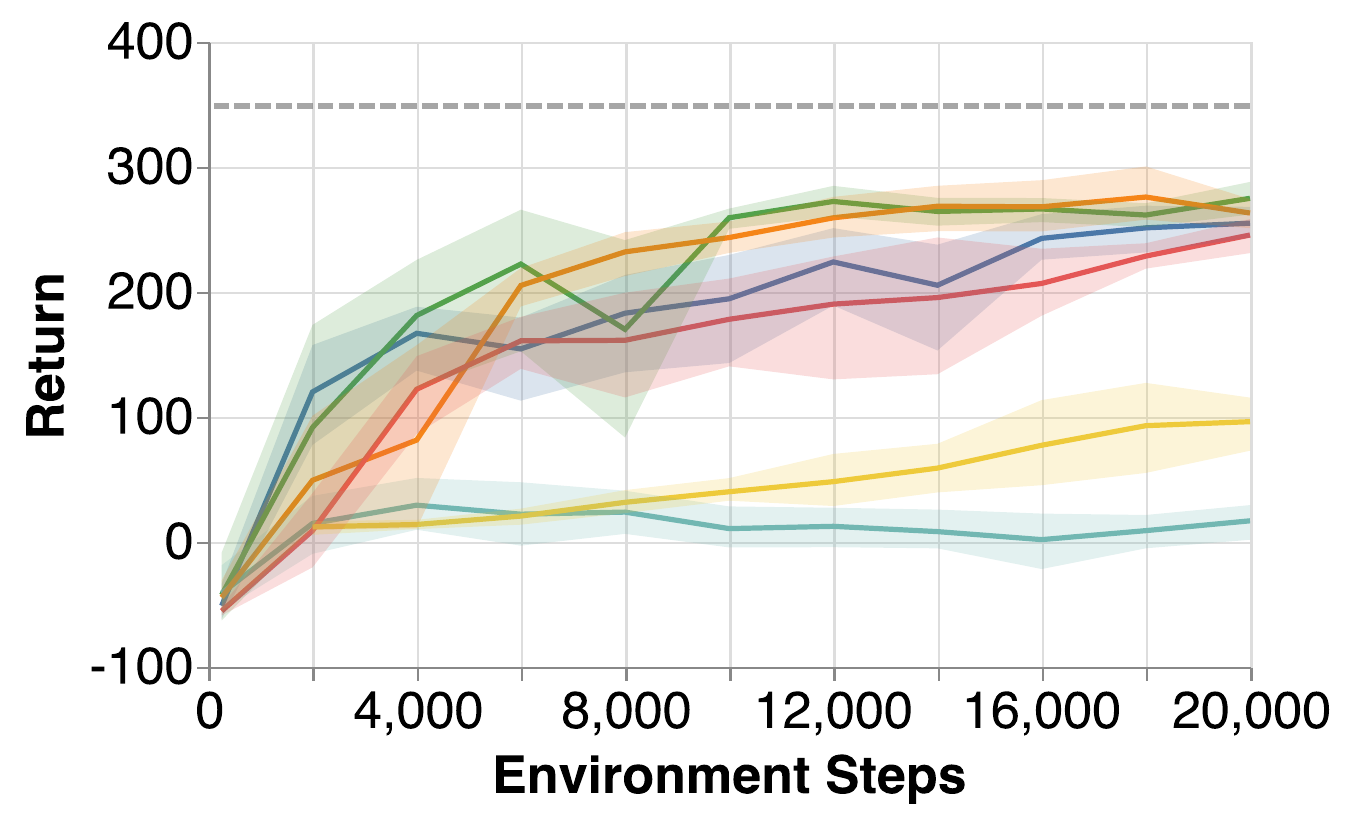}
        \caption{Running task performance}\label{fig:hc_main_a}
    \end{subfigure}
    \hfill
    \begin{subfigure}[b]{0.325\textwidth}
        \centering
        \includegraphics[width=\textwidth]{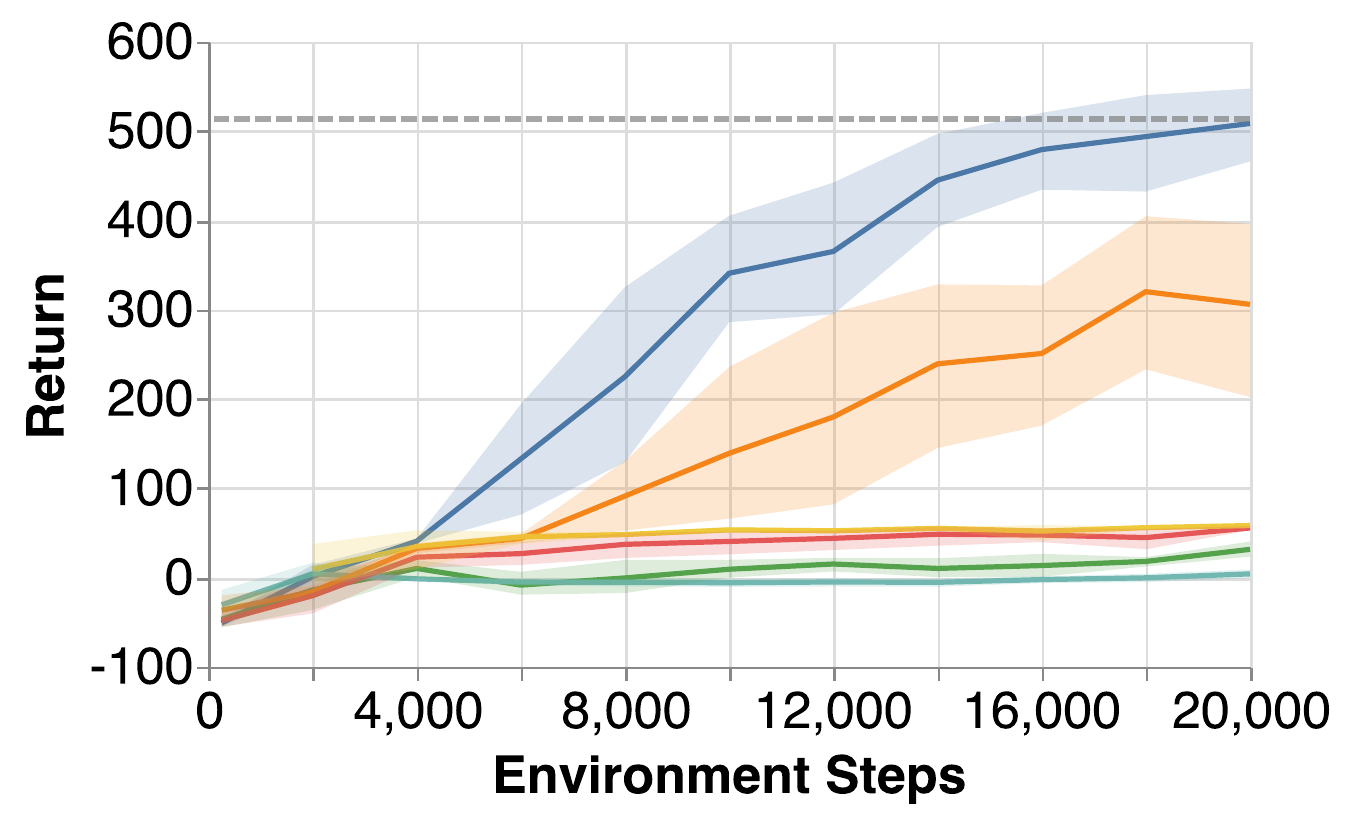}
        \caption{Flipping task performance}\label{fig:hc_main_b}
    \end{subfigure}
    \hfill
    \begin{subfigure}[b]{0.325\textwidth}
        \centering
        \includegraphics[width=\textwidth]{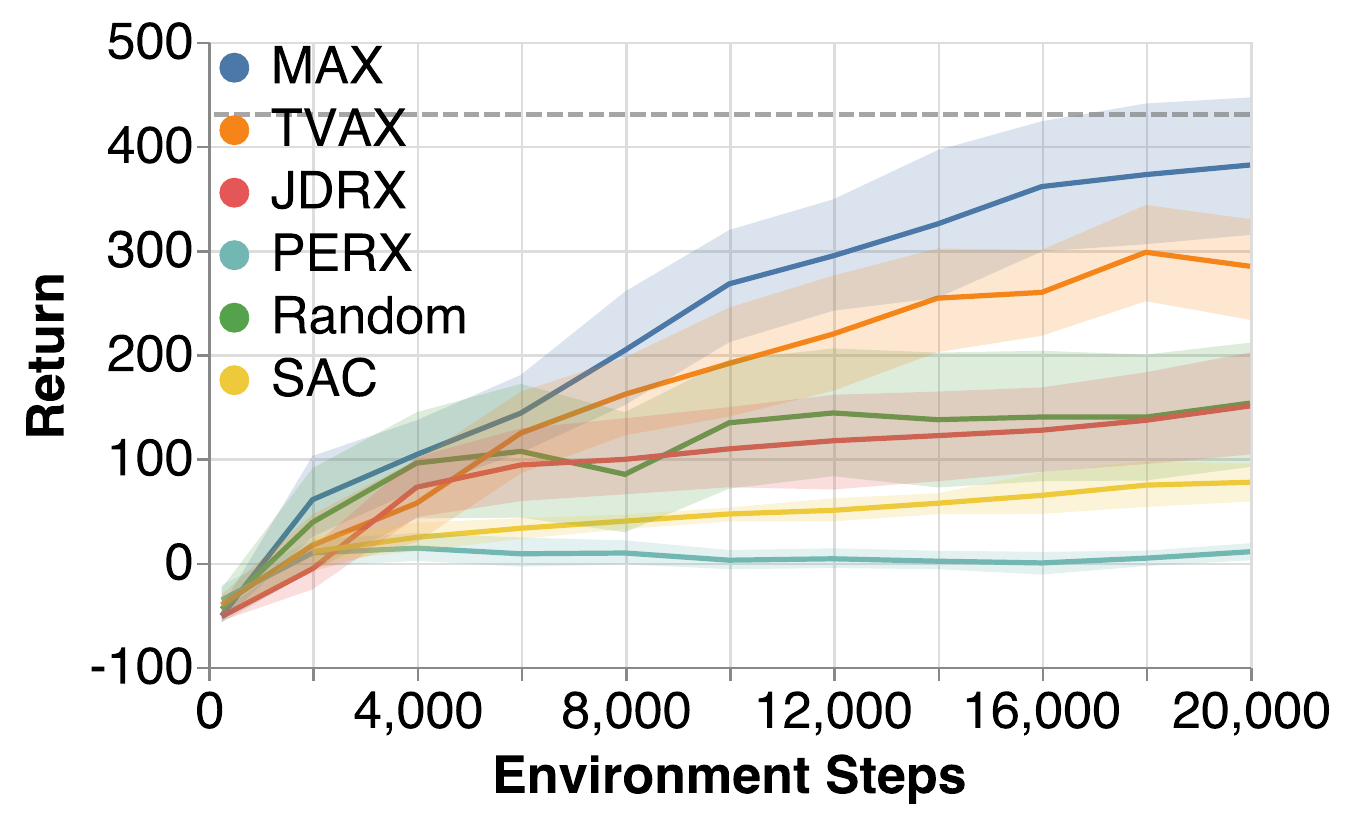}
        \caption{Average performance}\label{fig:hc_main_c}
    \end{subfigure}
    \caption{{\bf MAX on Half Cheetah tasks}.  The grey dashed horizontal line shows the average performance of an oracle model-free policy trained for $200k$ (10x more) steps by SAC in the environment, directly using the corresponding task-specific reward function. Notice that exploring the dynamics for the flipping task is more difficult than the running task as evidenced by the performance of the random baseline. Overall, active methods are quicker and better explorers compared to the reactive ones in this task. Each curve is the mean of $8$ runs.}
    \label{fig:hc_main}
\end{figure*}

\subsection{Continuous Environments}
To evaluate MAX in the high-dimensional continuous setting, two environments based on MuJoCo~\citep{todorov2012mujoco}, Ant Maze and Half Cheetah, were considered. The exploration performance was measured directly for Ant Maze, and indirectly in the case of Half Cheetah.

MAX was compared to four other exploration methods that lack at least one feature of MAX:
\vspace*{-3mm}
\begin{enumerate}
    \item \textbf{Trajectory Variance Active Exploration (TVAX):} an active exploration method that defines transition utilities as per-timestep variance in sampled trajectories in contrast to the per-state JSD between next state-predictions used in MAX.
    \item \textbf{Jensen-R\'enyi Divergence Reactive Exploration (JDRX):} a reactive counter-part of MAX, which learns the exploration policy directly from the experience collected so far without planning in the exploration MDP.
    \item \textbf{Prediction Error Reactive Exploration (PERX):} a commonly used reactive exploration method (e.g.\ in~\citet{pathakICMl17curiosity}), which uses the mean prediction error of the ensemble as transition utility.
    \item \textbf{Random} exploration policy.
\end{enumerate}

\begin{figure*}
    \centering
    \begin{subfigure}[b]{0.22\textwidth}
        \centering
        \includegraphics[width=\columnwidth]{{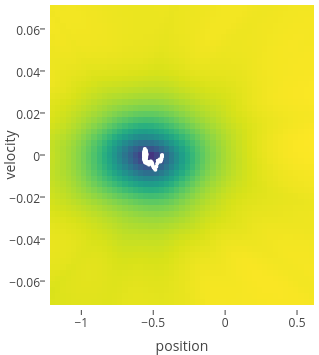}}
        \caption{100 steps\label{fig:mountain_car_100}}
    \end{subfigure}
    \hfill
    \begin{subfigure}[b]{0.22\textwidth}
        \centering
        \includegraphics[width=\textwidth]{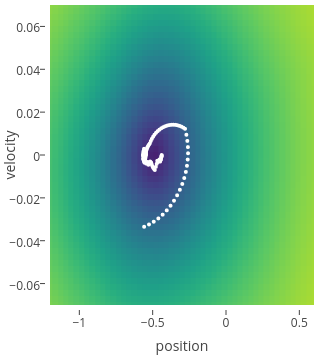}
        \caption{150 steps\label{fig:mountain_car_150}}
    \end{subfigure}
    \hfill
    \begin{subfigure}[b]{0.22\textwidth}
        \centering
        \includegraphics[width=\textwidth]{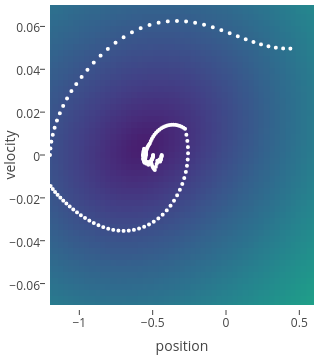}
        \caption{220 steps\label{fig:mountain_car_220}}
    \end{subfigure}
    \hfill
     \begin{subfigure}[b]{0.22\textwidth}
        \centering
        \includegraphics[width=\textwidth]{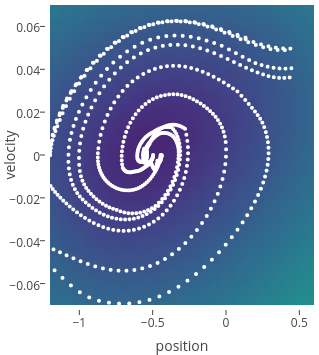}
        \caption{800 steps\label{fig:mountain_car_800}}
    \end{subfigure}
     \hfill
     \begin{subfigure}[b]{0.05\textwidth}
        \centering
        \includegraphics[width=\textwidth]{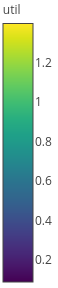}
        \vspace{0.5cm}
    \end{subfigure}
    \caption{{\bf Illustration of MAX exploration in the Continuous Mountain Car environment}. Each plot shows the state space of the agent, discretized as a 2D grid.  The color indicates the average uncertainty of a state over all actions. The dotted lines represent the trajectories of the agent. }
    \label{fig:mountain_car}
\end{figure*}

In the Ant Maze (see Figure~\ref{fig:mujoco_maze}), exploration performance was measured directly as the fraction of the U-shaped maze that the agent visited during exploration. In Half Cheetah, exploration performance was evaluated by measuring the usefulness 
of the learned model ensemble when exploiting it to perform two downstream tasks: running and flipping. For both environments, a Gaussian noise of $\cN(0, 0.02)$ was added to the states to introduce stochasticity in the dynamics. Appendix~\ref{app:continuous_env} details the setup.

Models were probabilistic Deep Neural Networks trained with negative log-likelihood loss to predict the next state distributions in the form of multivariate Gaussians with diagonal covariance matrices. Soft-Actor Critic (SAC;~\citealp{haarnoja2018soft}) was used to learn both pure exploration and task-specific policies.
The maximum entropy framework used in SAC is particularly well-suited to model-based RL as it uses an objective the both improves policy robustness, which hinders adversarial model exploitation, and yields multi-modal policies, which could mitigate the negative effects of planning with inaccurate models. 

Exploration policies were regularly trained with SAC from scratch with the
utilities re-calculated using the latest models to avoid over-commitment. The first phase of training involved only a fixed dataset containing the
transitions experienced by the agent so-far (agent history $D$). For the active methods (MAX and TVAX), this was followed by an additional phase where the
policies were updated by the data generated {\em exclusively} from the ``imaginary'' exploration MDP, which is the key feature distinguishing active from reactive exploration.

The results for Ant Maze and Half Cheetah are presented in Figures~\ref{fig:ant_maze_main}~and~\ref{fig:hc_main}, respectively. For Half Cheetah, an additional baseline, obtained by training an agent with model-free SAC using the task-specific reward in the environment, 
is included.
In both cases, the active exploration methods (MAX and TVAX) outperform the reactive ones (JDRX and PERX). Due to the noisy dynamics, PERX performs poorly. Among the two active methods, MAX is noticeably better as it uses a principled trajectory sampling and utility evaluation technique as opposed to the TVAX baseline which cannot distinguish the risk from the uncertainty. It is important to notice that the running task is easy since random exploration is sufficient
which is not the case for flipping where good performance requires directed active exploration.

None of the methods were able to learn task-oriented planning in Ant Maze, even with larger models and longer training times than were used to obtain the reported results. The Ant Maze environment is more complex than Half Cheetah and obtaining good performance
in downstream tasks using only the learned models is difficult due to other confounding factors such as compounding errors that arise in long-horizon planning. Hence, exploration performance was measured simply as the fraction of the maze the that agent explored. The inferior results of the baseline methods suggest that this task is non-trivial.

The evolution of the uncertainty landscape over the state-space of the environment when MAX is employed is visualized in
Figure~\ref{fig:mountain_car} for the Continuous Mountain Car environment. In the first exploration episode, the agent takes a spiral path through the state space (Figure~\ref{fig:mountain_car_220}): MAX was able to drive the car up and down the sides of valley to develop enough velocity to reach the mountain top
\textit{without any} external reward. In subsequent episodes, it carves out more spiral paths in-between the previous ones (Figure~\ref{fig:mountain_car_800}).

\section{Discussion}
An agent is \textit{meta-stable} if it is sub-optimal and, at the same time, has a policy that prevents it from gaining experience necessary to improve
itself~\citep{watkins1989learning}. 
Simply put, a policy can get stuck in a
local optimum and not be able to get out of it. In the simple cases,
undirected exploration techniques~\citep{thrun1992efficient} such as
adding random noise to the actions of the policy might be sufficient
to break out of meta-stability. If the environment is ergodic, then
reactive strategies that use exploration bonuses can solve
meta-stability. But active exploration of the form presented in this
paper, can in principle break free of any type of meta-stability.

Model-based RL promises to be significantly more
efficient and more general compared to model-free RL methods. However,
it suffers from model-bias~\citep{deisenroth2011pilco}: in certain
regions of the state space, the models could deviate significantly
from the external MDP. Model-bias can have many causes such
as improper generalization or poor exploration. A strong policy
search method could then exploit such degeneracy resulting in
over-optimistic policies that fail in the environment. Thorough
exploration is one way to potentially mitigate this issue. If learning certain aspects of the environment is difficult, it will manifest itself as disagreement in the ensemble. MAX would collect
more data about those aspects to improve the quality of models, thereby limiting adversarial exploitation by the policy. Since
model-based RL does not have an inherent mechanism to explore, MAX could be considered as an important addition to the model-based RL
framework rather than merely being an application of it.

\paragraph{Limitations.} The derivation in Section~\ref{sec:main} makes the assumption that the utility of a policy is the average utility of the probable transitions when the policy is used. However, encountering a subset of those transitions and training the models can change the utility of the remaining transitions, thereby affecting the utility of the policy. This second-order effect was not considered in the derivation. In the Chain environment 
for example, this effect leads to the agent planning to loop between pairs of uncertain states, rather than visiting many different uncertain states. MAX is also less computationally efficient in comparison to the baselines used in the paper as it trades off computational efficiency for data efficiency as is common in model-based algorithms.

\section{Related Work}
Our work is inspired by the framework developed in~\citet{Schmidhuber:97interesting,Schmidhuber:02predictable},
in which two adversarial reward-maximizing modules called the {\em left brain} and the {\em right brain} bet
on outcomes of experimental protocols or algorithms they have
collectively generated and agreed upon.  Each brain is intrinsically
motivated to outwit or surprise the other by proposing an experiment
such that the other {\em agrees} on the experimental protocol but {\em
  disagrees} on the predicted outcome.  After having
executed the action sequence protocol approved by both brains, the
surprised loser pays a reward to the winner in a zero-sum game. MAX
greatly simplifies this previous active exploration framework, distilling certain
essential aspects. Two or more predictive models that may compute different hypotheses
about the consequences of the actions of the agent, given
observations, are still used.  However, there is only one reward maximizer or
RL machine which is separate from the predictive models.

The information provided by an experiment was first analytically
measured by~\citet{lindley1956measure} in the form of expected
information gain in the Shannon
sense~\citep{Shannon:48}. \citet{bookFedorov} also proposed a theory
of optimal resource allocation during experimentation. By the 1990s,
information gain was used as an intrinsic reward for reinforcement
learning systems~\citep{storck1995reinforcement}. Even earlier,
intrinsic reward signals were based on prediction errors of a
predictive model~\citep{Schmidhuber:90sab} and on the learning
progress of a predictive model~\citep{Schmidhuber:91singaporecur}.  \citet{thrun1992efficient}
introduced notions of directed and undirected exploration in RL. 
Optimal Bayesian experimental
design~\citep{chaloner1995bayesian} is a framework for efficiently
performing sequential experiments that uncover a phenomenon. However,
usually the approach is restricted to linear models with Gaussian
assumptions. \citet{busetto2009optimized} proposed an optimal
experimental design framework for model selection of nonlinear
biochemical systems using expected information gain where they solve
for the posterior using the Fokker-Planck equation. In Model-Based Interval
Estimation~\citep{wiering1999explorations}, the uncertainty in the
transition function captured by a surrogate model is used to boost
Q-values of actions. In the context of
Active Learning, \citet{mccallumzy1998employing} proposed using
Jensen-Shannon Divergence between predictions of a committee of
classifiers to identify the most useful sample to be labelled next
among a pool of unlabelled samples. \cite{singh2005intrinsically} developed an intrinsic motivation framework inspired by neuroscience using prediction errors. \cite{itti2009bayesian} presented
the surprise formulation used in Equation~\ref{eq:kl} and demonstrated
a strong correlation between surprise and human
attention. At a high-level, MAX can be seen as a form of Bayesian Optimization~\citep{snoek2012practical} adopted for exploration in RL which employs an inner search-based optimization during planning. Curiosity has also been studied extensively from the perspective of developmental robotics~\citep{oudeyer2018computational}. \citet{Schmidhuber:09abials} suggested a general form of learning progress as compression progress which can be used as an extra intrinsic reward for curious RL systems.

Following these, \citet{sun2011planning} developed an optimal Bayesian framework for
curiosity-driven exploration using learning progress. After proving
that Information Gain is additive in expectation, a dynamic
programming-based algorithm was proposed to maximize Information
Gain. Experiments however were limited to small tabular MDPs
with a Dirichlet prior on transition
probabilities. A similar Bayesian-inspired, hypothesis-resolving, model-based RL exploration algorithm was proposed in~\citet{hester2012intrinsically} and shown to outperform prediction error-based and other intrinsic motivation methods. In contrast to MAX, planning uses the mean prediction of a model ensemble to optimize a disagreement-based utility measure which is augmented with an additional state-distance bonus.
\citet{still2012information} derived a 
exploration and exploitation trade-off in an attempt to maximize the predictive
power of the agent. \citet{mohamed2015variational} combined Variational
Inference and Deep Learning to form an objective based on mutual
information to approximate agent empowerment. In comparison to our
method, \citet{houthooft2016vime} presented a Bayesian approach to
evaluate the value of experience taking a reactive
approach. However, they also used Bayesian Neural Networks to maintain
a belief over environment dynamics, and the information gain to
bias the policy search with an intrinsic reward
component. Variational Inference was used to approximate the
prior-posterior KL divergence. \citet{bellemare2016unifying} derived a
notion of pseudo-count for estimating state visitation frequency in
high-dimensional spaces. They then transformed this into a form of
exploration bonus that is maximized using DQN. \citet{osband2016deep}
propose Bootstrapped DQN which was used as a
baseline. \citet{pathakICMl17curiosity} used inverse models to avoid learning anything that the agent cannot control to reduce risk, and prediction error in the latent space to perform reactive exploration. A large-scale study
of curiosity-driven exploration \citep{burda2018large} found that curiosity
is correlated with the actual objectives of many environments, and
reported that using random features mitigates some of the
non-stationarity implicit in methods based on curiosity. \citet{eysenbach2018diversity} demonstrated the power of optimizing policy diversity in the absence of a reward function in developing skills which could then be exploited.

Model-based RL has long been touted as the cure for sample
inefficiency problems of modern
RL~\citep{Schmidhuber:90diffgenau,sutton1991reinforcement,deisenroth2011pilco}. Yet learning
accurate models of high-dimensional environments and exploiting them
appropriately in downstream tasks is still an active area of
research. Recently, \citet{kurutach2018model} and \citet{chua2018deep}
have shown the potential of model-based RL when combined with Deep
Learning in high-dimensional environments. In particular, this work was
inspired by~\citet{chua2018deep} who combined probabilistic models with novel trajectory sampling techniques using particles
to obtain better approximations of the returns in the environment. Concurrent with this work, \citet{pathakICMl19disagreement} also showed the advantages of using an ensemble of models for exploring complex high-dimensional environments, including a real robot.

\section{Conclusion}
This paper introduced MAX, a model-based RL algorithm for pure exploration. It
can distinguish between learnable and unlearnable unknowns and search
for policies that actively seek learnable unknowns in the
environment. MAX provides the means to
use an ensemble of models for simulation and evaluation of an exploration policy. The quality of the exploration policy can therefore be
directly optimized without actual interaction with the environment. Experiments
in hard-to-explore discrete and high-dimensional continuous environments indicate that MAX
is a powerful generic exploration method.

\section*{Acknowledgements}
We would like to thank  J\"{u}rgen Schmidhuber, Jan Koutn\'{i}k, Garrett Andersen, Christian Osendorfer, Timon Willi, Bas Steunebrink, Simone Pozzoli, Nihat Engin Toklu, Rupesh Kumar Srivastava and Mirek Strupl for their assistance and everyone at NNAISENSE for being part of a conducive research environment.

\bibliography{example_paper}
\bibliographystyle{icml2019}

\appendix

\onecolumn 

\section{Policy Evaluation}
\label{sec:appendix_policy_eval}
Let $\mathrm{IG}(\pi)$ be the information gain of a policy $\pi$:
\begin{align}   
    \mathrm{IG}(\pi) &= \Eb{\phi \sim \cP(\Phi \given \pi)}{\mathrm{IG}(\phi)}, \nonumber \\
    &= \int_{\Phi} \mathrm{IG}(\phi) p(\phi \given \pi) \, d\phi, \label{eq:ig}
\end{align}
where $p(\phi \given \pi) = p(s, a, s' \given \pi)$ is the probability of transition $(s, a, s')$ occurring given $\pi$ is the behavioural policy in the external MDP.

(\ref{eq:ig}) can be expanded as
\begin{align}
    \mathrm{IG}(\pi) &= \int_{\cS} \int_{\cA} \int_{\cS} \mathrm{IG}(s, a, s') p(s, a, s' \given  \pi)  \, ds' \, da \, ds \nonumber \\
    &= \int_{\cS} \int_{\cA} \int_{\cS} \mathrm{IG}(s, a, s') p(s' \given s, a) p(s, a \given \pi) \, ds' \, da \, ds \nonumber \\
    &= \int_{\cS} \int_{\cA} u(s, a) p(s, a \given \pi)  \, da \, ds, \nonumber \\
    &= \int_T \int_{\cS} \int_{\cA} u(s, a) p(s, a \given \pi, t) p(t)\, da \, ds \, dt \label{eq:expanded_pi_utility}
\end{align}
where $u(s, a)$ is action utility function
\begin{equation}
    \label{eq:action-utility-def}
    u(s, a) = \int_{\cS} \mathrm{IG}(s, a, s') p(s' \given s, a) \, ds',
\end{equation}
quantifing the net utility of taking action $a$ from state $s$.

Equation~\ref{eq:expanded_pi_utility} reduces then to the following expectation:
\begin{equation}
    \label{eq:policy-utility}
    \mathrm{IG}(\pi) = \Eb{t \sim \cP(T)}{\Eb{s, a \sim \cP(\cS, \cA \given \pi, t)} {u(s, a)}}
\end{equation}

\section{Action Evaluation}
\label{sec:appendix_action_eval}
To obtain a closed form expression for $u(s, a)$, Equation~\ref{eq:action-utility-def} can be expanded using Equation~\ref{eq:kl} to:
\begin{align*}
    u(s, a) &= \int_{\cS} \mathrm{IG}(s, a, s') p(s' \given a, s) \, ds', \\
    &= \int_{\cS} \mathcolor{red}{\kl{\cP(T \given \phi)}{\cP(T)}} \: p(s' \given s, a) \, ds'\\
\end{align*}
Expanding from definition of KL-divergence:
\[
\mathcolor{red}{\kl{\cP_1}{\cP_2}} = \mathcolor{blue}{\int_Z p_1(z) \log{\left(\frac{p_1(z)}{p_2(z)}\right)} \, dz}
\]
we get
\begin{equation}
    \label{eq-pre-bayes}
    u(s, a) = \int_{\cS} \mathcolor{blue}{\int_T p(t \given \phi) \log{\Bigg(\frac{p(t \given  \phi)}{p(t)}\Bigg)}} \: p(s' \given s, a) \, \mathcolor{blue}{dt} \, ds'
\end{equation}
Using the Bayes rule
\begin{equation}
    \label{eq-bayes}
    p(t \given s', a, s) = \frac{p(s' \given s, a, t) p(t \given s,a)}{p(s' \given s,a)}
\end{equation}
and the observation that $p(t \given s, a) = p(t)$ \cite{sun2011planning} leads to:
\begin{align}
    u(s, a) &= \int_{\cS} \int_{T} \frac{p(s' \given s,a,t) p(t)}{p(s' \given s,a)} \log{\Bigg(\frac{p(s' \given s,a,t)}{p(s'|s,a)}\Bigg)} \: p(s' \given s, a) \, dt \,  ds' \nonumber\\
    &= \int_{\cS}  \int_T p(s' \given  s, a, t) p(t )\log{\Bigg(\frac{p(s' \given  s, a, t)}{p(s' \given  s, a, \bar{t})}\Bigg)}  \, dt \,  ds'. \label{eq:pre-log-expansion}
\end{align}

Expanding and swapping integrals in the former term:
\begin{equation}
\label{eq:pre-entropy}
\begin{split}
u(s, a) &= \int_T \int_{\cS} \mathcolor{red}{p(s' \given  s, a, t)} \mathcolor{blue}{\log{(p(s' \given  s, a, t))}}p(t )  \,  ds' \, dt \\
& \quad - \quad \int_{\cS}  \mathcolor{red}{\int_T p(s' \given  s, a, t) p(t )}\,  \mathcolor{red}{dt} \mathcolor{blue}{\log{(\int_T p(s' \given  s, a, t)p(t ) \, dt}}) \, ds'
\end{split}
\end{equation}

$ - \int_z \mathcolor{red}{p(z) \mathcolor{blue}{\log{(p(z))}}} \, dz$ is just entropy $\mathfrak{H}(\cP(z))$.
\begin{align}
    u(s, a) &= \mathcolor{green}{\mathfrak{H}\left(\int_T \cP(\cS \given  s, a, t)p(t ) \, dt \right)} - \mathcolor{orange}{\int_T \mathfrak{H}(\cP(\cS \given  s, a, t))p(t ) \, dt}\\
    &= \mathfrak{H}\left(\Eb{t \sim \cP(T )}{ \cP(\cS \given  s, a, t)} \right) - \Eb{t \sim \cP(T )}{ \mathfrak{H}(\cP(\cS \given  s, a, t))}
\end{align}

$\cP(\cS \given  s, a, t)$ represents a probability distribution over the next state with removal of the integrals over $s'$. Given a space of distributions, the entropy of the average distribution minus the average entropy of distributions is the Jensen Shannon Divergence (JSD) of the space of distributions. It is also termed as the Information Radius and is defined as:
\begin{equation}
    \label{eq-jsd}
    \mathrm{JSD}\{\cP_{\bar{Z}}  \, \mid \, \cP_{\bar{Z}} \,  \sim  \, \cP(\mathbf{Z})\} = \mathcolor{green}{\mathfrak{H}\left(\int_{\bar{Z}} \cP_{\bar{Z}} p(\cP_{\bar{Z}}) \, d{\bar{Z}}\right)} - \mathcolor{orange}{\int_{\bar{Z}} \mathfrak{H}(\cP_{\bar{Z}}) p(\cP_{\bar{Z}}) \, d{\bar{Z}}}
\end{equation}
where $\mathbf{Z}$ is the space of distributions $\cP_{\bar{Z}}$.  $\cP(\mathbf{Z})$ is a compound probability distribution with $p(\cP_{\bar{Z}})$ as its density function.

Therefore, Equation~\ref{eq:action-utility-def} can be expressed as:
\begin{equation}
    u(s, a) = \mathrm{JSD}\{\cP(\cS \given  s, a, t)  \, \mid \,   \, t \,  \sim  \, \cP(T )\}
\end{equation}
where $T$ is the space of transition functions with probabilistic outputs.

\section{Chain Environment\label{app:chain_env}}
\subsection{Stochastic Environment}
The left-most state (state $0$) of the Chain
Environment was modified to be a \textit{stochastic trap} state. If
the agent is in the trap state, regardless of the chosen action, it is
equally likely to remain in state $0$ or end up in state $1$. A method
that relies only on prediction errors cannot separate risk from
uncertainty. Hence it would repeatedly visit the trap state as it is
easier to reach it compared to a far-off unexplored state. Figure~\ref{fig:stochastic_ablation} compares the performance
of our method on both the normal Chain environment and the one
with a stochastic trap state. Although MAX is slowed down, it still
manages to explore the transitions. The stochastic trap state
increases the variance in the output of the models. This is because the models
were trained on different outputs for the same input. Hence the
resulting output distribution of the models were sensitive to the
training samples the models have seen. This can cause
disagreements and hence result in a higher utility right next to the
initial state. The true uncertainty on the right, which is fainter, is
no longer the dominant source of uncertainty. This causes even our
method to struggle, despite being able to separate the two forms of
uncertainty.

\subsection{MAX}
The ensemble consisted of $3$ forward models implemented as
fully-connected neural networks, each receiving one-hot-encoded state
of the environment and the action of the agent as input. The outputs
of the network were categorical distributions over the (next)
states. The networks were independently and randomly initialized which
was found to be enough to foster the diversity for the
out-of-sample predictions and hence bootstrap sampling was not used.
After 3-episode warm-up with random actions, the models were trained for $150$ iterations. Searching
for an exploration policy in the exploration MDP, which was an
open-loop plan, was performed using $25$ rounds of Monte Carlo Tree
Search (MCTS), which used $5$ random trajectories for each step of
expansion with Thompson Sampling as the selection policy. The best
action was chosen based on the average value of the children. Models
were trained after experiencing each transition in the environment and
the MCTS tree was discarded after every step. The hyper-parameters of MAX
were tuned with grid search for the chain length of $N=50$.

\begin{table}[H]
  \caption{Hyper-Parameters for MAX. \textit{Grid Size: 0.8k}}
  \label{hyp-table:ours}
  \begin{center}
    \begin{tabular}{lcccc}
      \toprule
      {\bf Hyper-parameter} & \multicolumn{4}{c}{{\bf Values}} \\
      \midrule
      Hidden Layers & $2$ 	& $3$ & \textbf{$4$} & \\
      Hidden Layer Size 		  & $64$ 	& \textbf{$128$} & $256$ & \\
      Learning Rate           & \textbf{$10^{-3}$} & $10^{-4}$ & & \\
      Batch Size              & $16$ & $64$ & \textbf{$256$} & \\
      Training Iterations per Episode     & $16$ & $32$ &       \textbf{$64$} & $128$\\
      Weight Decay            & $0$ & $10^{-5}$ & \textbf{$10^{-6}$}   &     $10^{-7}$\\
      \bottomrule
    \end{tabular}
  \end{center}
\end{table}

\subsection{Baselines}
\textbf{Exploration Bonus DQN} is an extension of the DQN algorithm~\citep{mnih2015human}, in which the transitions that are visited for the first time carry an extra bonus reward. This causes the value of those transitions to be temporarily over-estimated and results in more visits to novel regions of the environment. In general, the exploration bonuses can be computed using prediction errors of forward models~\citep{pathakICMl17curiosity}. However, for our simple environment, the bonuses were provided by an oracle, implemented as a transition look-up table. This is equivalent to having an ideal model that can learn about a transition in one-shot, therefore emulating the best-case scenario for the method.

\textbf{Bootstrapped DQN} by ~\cite{osband2016deep} also extends the DQN algorithm and it is based on the same underlying principle as Exploration Bonus DQN. Instead of introducing arbitrary extra rewards, it relies on stochastic over-estimation of values in novel states. Multiple $Q$-value networks or \textit{heads} are maintained and trained on a shared replay buffer. Before every episode, a head is randomly selected and the agent acts greedily with respect to it. If a novel transition $(s, a, s')$ is added to the replay buffer, the hope is that at least one of the heads over-estimates the value of some action $a'$ in state $s$ or $s'$, causing the policy resulting from TD-bootstrapping to prefer this state. This leads to that particular state being further explored when that head is used to act.

\begin{table}[H]
  \caption{Hyper-Parameters for DQN with Exploration Bonus.   \textit{Grid Size: 4.3k}}
  \label{hyp-table:eb-dqn}
  \begin{center}
    \begin{tabular}{lccc}
      \toprule
      {\bf Hyper-parameter} & \multicolumn{3}{c}{\bf Values} \\
      \midrule
      Exploration Bonus   & $5 \times 10^{-3}$ & $10^{-2}$ &       \textbf{$10^{-1}$} \\
      Replay Size     & \textbf{$256$}& $10^{5}$&\\
      Hidden Layers    & $1$& $2$& \textbf{$3$} \\
      Hidden Layer Size & $32$& \textbf{$64$}& $128$\\
      Learning Rate    & \textbf{$10^{-2}$}& $10^{-3}$& $10^{-4}$ \\
      Batch Size    & $16$& \textbf{$32$}& $64$ \\
      Discount Factor      & $0.95$& \textbf{$0.975$}& $0.99$ \\
      Target Network Update Frequency      & $16$& $64$&     \textbf{$256$}           \\
      \bottomrule
    \end{tabular}
  \end{center}
\end{table}

\begin{table}[H]
  \caption{Hyper-Parameters for Bootstrapped DQN. \textit{Grid Size:   2.1k}}
  \label{hyp-table:boot-dqn}
  \begin{center}
    \begin{tabular}{lccc}
      \toprule
      {\bf Hyper-parameter} & \multicolumn{3}{c}{{\bf Values}} \\
      \midrule
      Hidden Layers    & \textbf{$1$} & $2$ & $3$ \\
      Hidden Layer Size & \textbf{$32$} & $64$ & $128$\\
      Learning Rate    & $10^{-2}$ & $10^{-3}$ & \textbf{$10^{-4}$}   \\
      Batch Size    & $16$ & \textbf{$32$} & $64$ \\
      Training Iterations per Episode     & $16$ & $32$ &       \textbf{$64$}\\
      Discount Factor      &\textbf{$0.95$} & $0.975$ & $0.99$ \\
      Target Network Update Frequency      & \textbf{$16$} & $64$ &       $256$\\
      \bottomrule
    \end{tabular}
  \end{center}
\end{table}

Commonly used $\epsilon$-greedy exploration was turned off in both baselines, thereby making exploration reliant solely on the methods themselves. The $Q$-value functions were implemented with multi-layer fully connected neural networks. The state was encoded with thermometer encoding for Bootstrapped DQN as proposed by~\citet{osband2016deep}. The $Q$-value network in Exploration Bonus DQN was trained with one batch of data after each step in the environment. For Bootstrapped DQN, $10$ $Q$-value network heads were trained \textit{only after} each episode, with the exact number of iterations being tuned. Bootstrap sampling was not employed and all heads were trained on all transitions collected so far. Target networks and replays were used for both methods as in the standard DQN. RMSprop optimizer with a momentum of $0.9$ was used to minimize the Huber Loss with gradient clipping of 5.

All neural networks consisted of Glorot-initialized, \textit{tanh}-activated fully connected layers with their depth and width being tuned.

Each hyper-parameter configuration was tested using $5$ different random seeds. Hyper-parameters were ranked according to the median of the area under the exploration curve.

\subsection{Supplementary Figures}
\begin{figure}[H]
    \begin{subfigure}[b]{0.49\textwidth}
        \includegraphics[height=150pt]{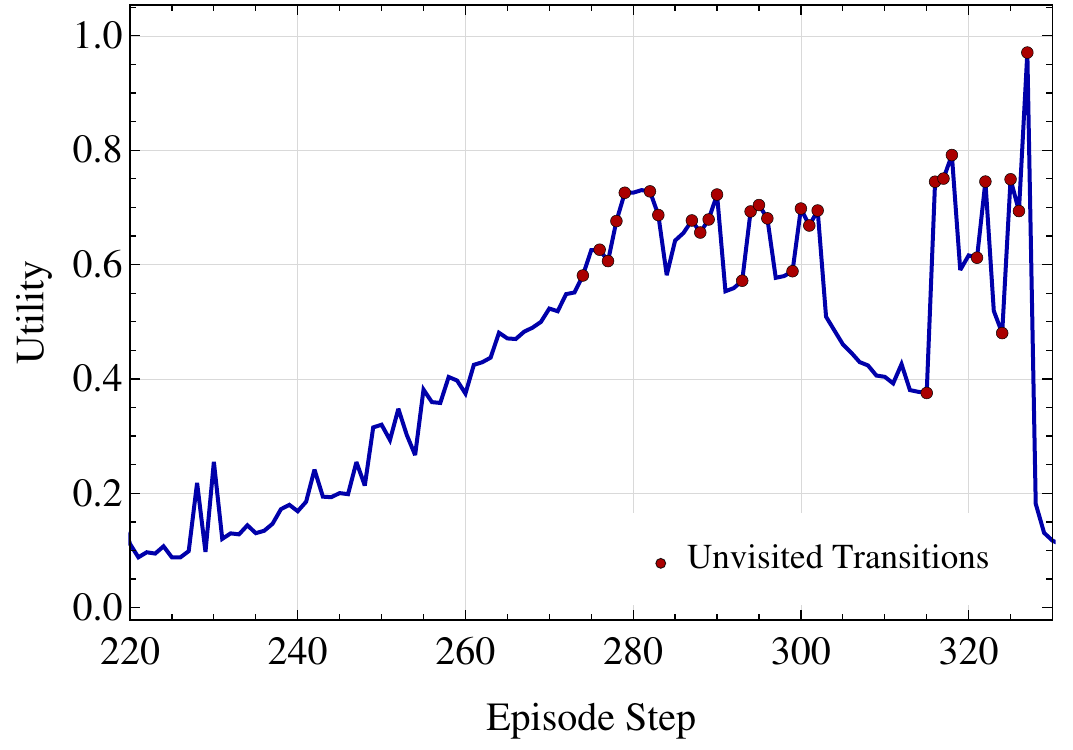}
        \caption{\label{fig:pig_over_time}}
    \end{subfigure}
    \hfill
    \begin{subfigure}[b]{0.49\textwidth}
        \includegraphics[height=150pt]{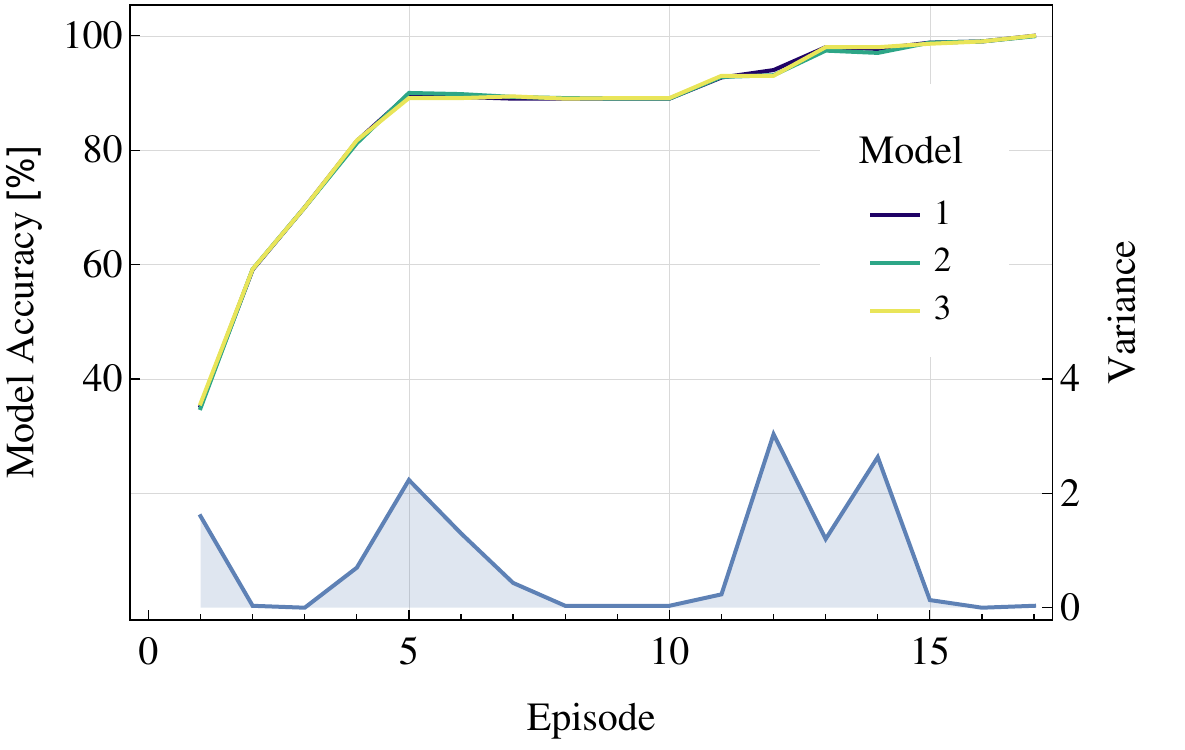}
        \caption{\label{fig:model_learning}}
    \end{subfigure}
    \caption{Exploration with MAX. (\subref{fig:pig_over_time}) Utility of the exploration policy (in normalized log scale) during an exemplary episode. The red points mark encounters with novel transitions. (\subref{fig:model_learning}) The accuracy of the models in the ensemble at each episode and the variance of model accuracy.
    }
\end{figure}

\begin{figure}[H]
    \centering
    \includegraphics[width=\textwidth]{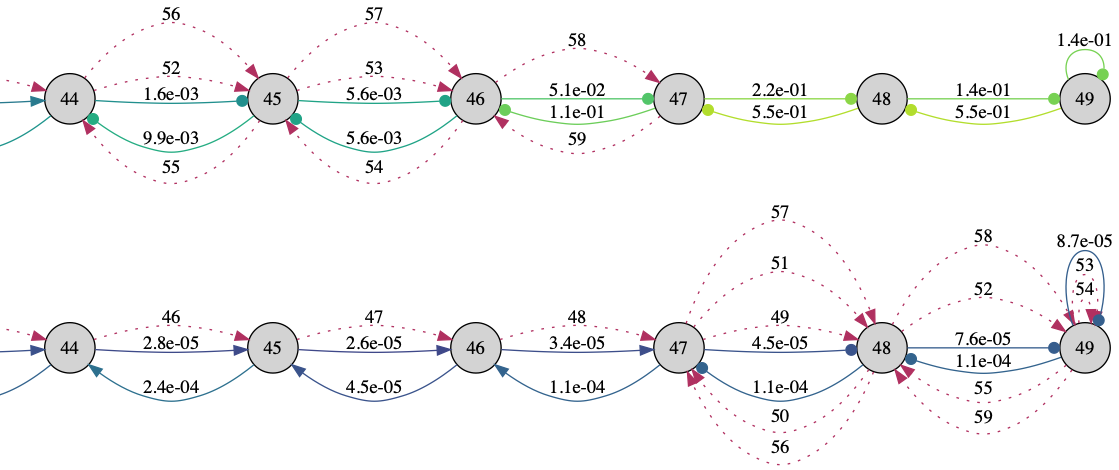}
    \caption{Instance of uncertainty minimization:MAX behaviour in the second-to-last (top) and the last (bottom) episodes. The solid arrows (environment transitions) and are annotated with the utilities of predicted information gains. The arrowheads are circular for the unvisited transitions and regular for the visited ones (before an episode). The dotted arrows with step numbers visualize the path the agent was following during an episode (only the final steps are shown).
    }
    \label{fig:uncertainty_minimization}
\end{figure}

Figures~\ref{fig:pig_over_time},
\ref{fig:model_learning} and~\ref{fig:uncertainty_minimization} analyze the behaviour of the proposed
algorithm. Fig.~\ref{fig:pig_over_time} plots the utility of the
exploration policy during two learning episodes. Clearly, high utility
correlates with experiencing novel
transitions. Figure~\ref{fig:model_learning}, which shows the learning
curves of individual models in an ensemble, indicates that less than
$20$ episodes is required to correctly learn all the transitions. The
surrogate models were therefore indistinguishable from the
environment. Notice also that the learning curves of the three models
are close to each other. This is because the models nearly always
agree on the transitions from the training set and disagree on the
others. The exploration is thus mainly driven by the divergence of
predictions for the unobserved
transitions. Figure~\ref{fig:uncertainty_minimization} visualizes the
transitions in the final two episodes of a run showing how MAX plans
for uncertain (unvisited) transitions, which results in minimizing
their corresponding uncertainties.

\begin{figure}[H]
    \centering
    \begin{subfigure}[b]{0.325\textwidth}
        \centering
        \includegraphics[height=120pt]{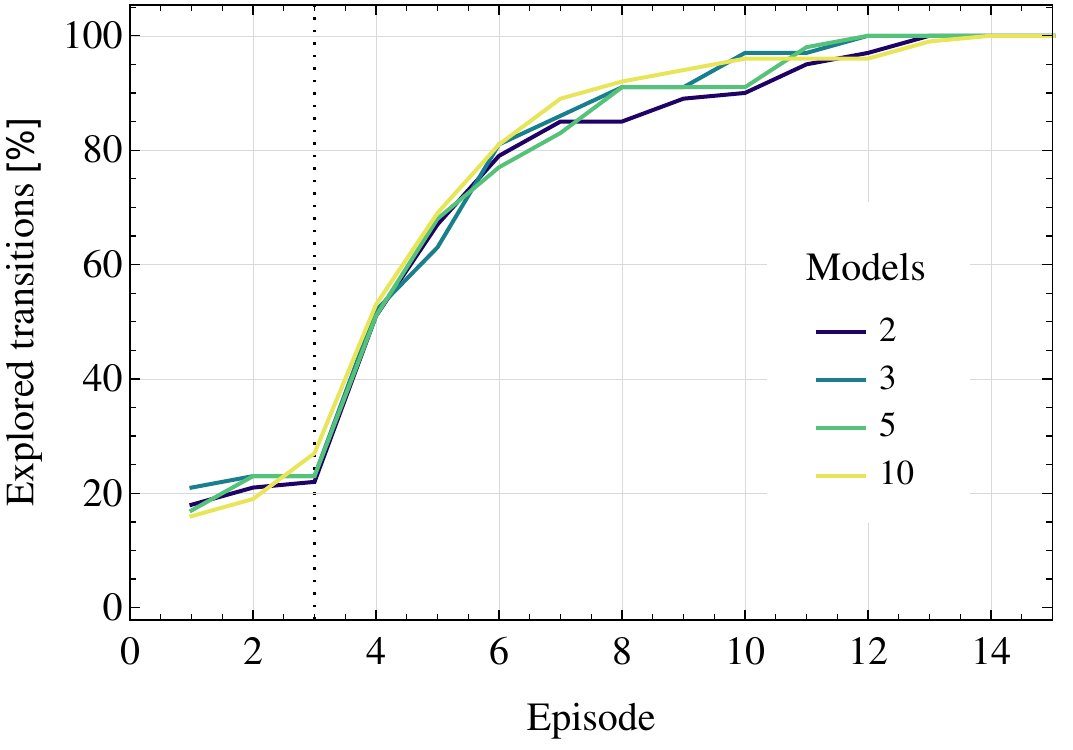}
        \caption{\label{fig:models_ablation}}
    \end{subfigure}
    \hspace*{1mm}
    \begin{subfigure}[b]{0.325\textwidth}
        \centering
   \hspace*{1mm}
        \includegraphics[height=120pt]{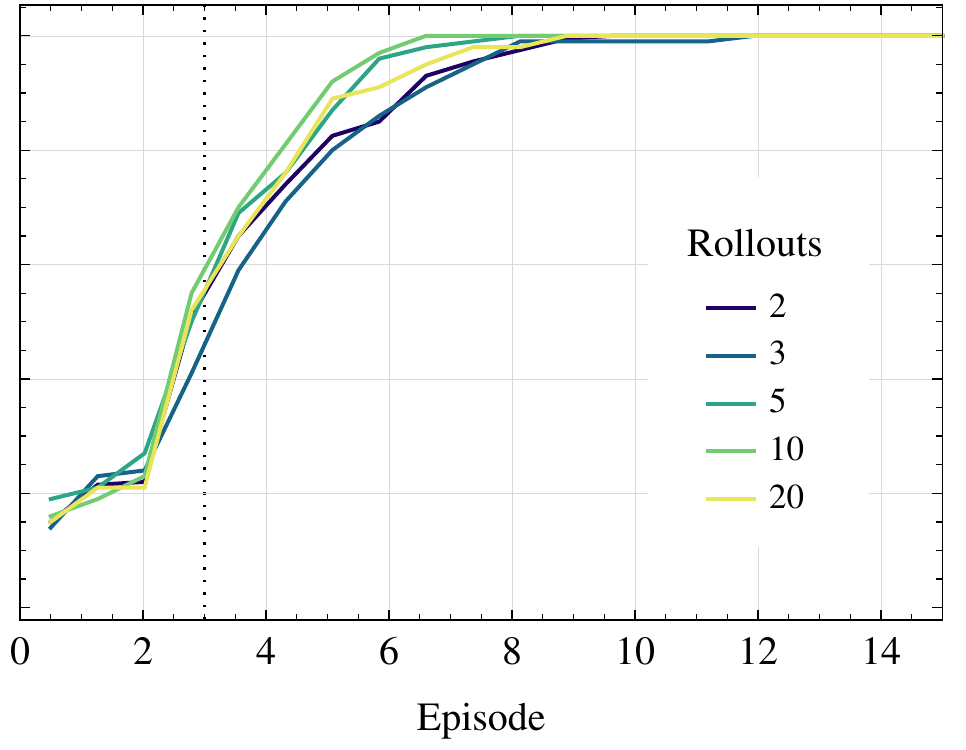}
       \caption{\label{fig:roll-out_ablation}}
    \end{subfigure}
    \begin{subfigure}[b]{0.325\textwidth}
        \centering
        \includegraphics[height=120pt]{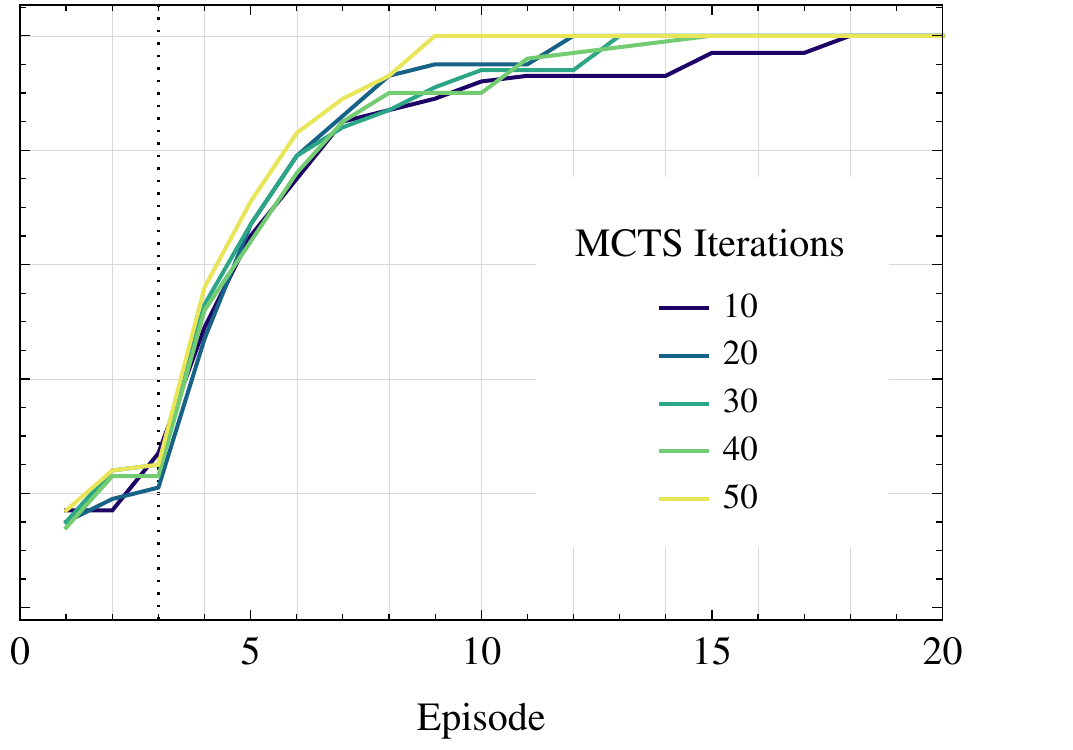}
        \caption{\label{fig:planning_ablation}}
    \end{subfigure}
    \caption{Algorithm Properties. Each learning curve is a median of
      $5$ runs with different seeds. Vertical dotted line marks the
      ends of the warm-up phase. Sub-figures show how the percentage
      of explored transitions varies with respect to
      (\subref{fig:models_ablation}) ensemble size,
      (\subref{fig:roll-out_ablation}) number of rollouts and
      (\subref{fig:planning_ablation}) planning iterations.}
    \label{fig:robustness}
\end{figure}

\paragraph{Robustness to Key Hyper-Parameters}
Number of models in the ensemble, trajectories per an MCTS iteration, and MCTS iterations were varied and the results are shown in Figure~\ref{fig:robustness}. In
general, more trajectories, more models in the ensemble and more
planning iterations lead to better performance. Notice, however, that
the algorithms works efficiently also for minimal settings.

\section{Continuous Environments\label{app:continuous_env}}

\subsection{Numerically Stable Jensen-R\'enyi Divergence Implementation}
\[
\mathfrak{D}(\cN_i, \cN_j) = \frac{1}{\left|\Sigma_i+\Sigma_j\right|^{\frac{1}{2}}} \exp \left(-\frac{1}{2}\left(\mu_j - \mu_i\right)^{T}\left(\Sigma_i+\Sigma_j\right)^{-1}\left(\mu_j - \mu_i\right)\right)
\]

\begin{align*}
    p &= \left(u_j-u_i\right)^{T}\left(\Sigma_i+\Sigma_j\right)^{-1}\left(\mu_j - \mu_i\right)\\
    q &= \ln(\left|\Sigma_i+\Sigma_j\right|)\\
    &= \trace(\ln(\Sigma_i + \Sigma_j))
\end{align*}
\begin{align*}
    \mathfrak{D}(\cN_i, \cN_j) &= \exp \left(-\frac{1}{2}(p + q)\right)
\end{align*}

Additionally, \textsc{log-sum-exp} trick is used to calculate the entropy of the mean.

\subsection{Experimental Setup}
256 steps of warm-up was used for all methods to train the models before exploration began. During exploration, for all methods, models were retrained from scratch for 50 epochs every 25 steps. SAC policies were also relearned from scratch every 25 steps to avoid over-commitment. The current exploration experience was transferred to SAC as its off-policy data and 100 steps of SAC off-policy training was done. For active methods, there was additional on-policy training done using data generated during 50 episodes, each of length 50, starting from current state using 128 actors and the model ensemble. The SAC entropy weight $\alpha$ was tuned separately for each method. It was found to be $0.02$ for R\'enyi entropy based methods MAX and JDRX, and $0.2$ for the TVAX and PERX.

Exploitation in Half Cheetah was done every 2000 steps of exploration. Using the exploration data, the model ensemble was trained for 200 epochs. For all methods, SAC policies were trained using both off-policy data from exploration and with on-policy data from the model ensemble. On-policy data was generated using 250 episodes of length 100 using 128 actors. Note that this is recursive prediction for 100 steps starting from the start state. As with exploration, the next states were sampled at each transition by randomly selecting a model from the ensemble and then sampling the next state from its output distribution. A policy was trained and evaluated 3 times and the average performance was reported.

Models were trained to predict the state deltas, instead of raw next states as is common. The states, actions and deltas were all normalized to have zero mean and unit variance. For all methods, the utility computation was done in normalized scales for stability. 

\textbf{Half Cheetah Reward Functions}\\
Running: $r_t = v^x_t - 0.1 \|a_t\|^2_2$; flipping: $r_t = \omega^y_t - 0.1 \|a_t\|^2_2,$

where $v^x_t$ is the velocity along the $x$ axis at time $t$, and $\omega^y_t$ is the angular velocity around axis $y$ at time time $t$.

\begin{table}
  \caption{Hyper-Parameters for Models in Continuous Environments}
  \label{hyp-table:continuous_envs}
  \vskip 0.2in
  \begin{center}
    \begin{tabular}{lc}
      \toprule
      {\bf Hyper-parameter} & {{\bf Value}} \\
      \midrule
      Ensemble Size & 32\\
      Hidden Layers    & 4 \\
      Hidden Layer Size & 512   \\
      Batch Size    & 256 \\
      Non-linearity & Swish \\
      Learning Rate & 0.001\\
      
      \bottomrule
    \end{tabular}
  \end{center}
\end{table}

\end{document}